\pdfoutput=1

\documentclass[11pt]{article}

\usepackage[final]{acl}

\usepackage{times}
\usepackage{latexsym}

\usepackage[T1]{fontenc}

\usepackage[utf8]{inputenc}

\usepackage{microtype}

\usepackage{inconsolata}
\usepackage{multirow}
\usepackage[capitalise, noabbrev]{cleveref}
\usepackage{booktabs}
\usepackage{graphicx}

\usepackage[normalem]{ulem} 
\usepackage{xcolor}
\usepackage{xspace}
\def\change#1{#1}

\title{Prompting LLMs: Length Control for Isometric Machine Translation}

\author{Dávid Javorský$^1$ \and Ondřej Bojar$^1$ \and François Yvon$^2$
  \\ \\
  $^{1}$Charles University, Faculty of Mathematics and Physics, Prague, Czechia\\
  $^2$Sorbonne Université, CNRS, ISIR, Paris, France\\
  \texttt{\{javorsky,bojar\}@ufal.mff.cuni.cz}~~~\texttt{francois.yvon@cnrs.fr}}

\begin{document}
\maketitle
\begin{abstract}
In this study, we explore the effectiveness of isometric machine translation across multiple language pairs (En$\to$De, En$\to$Fr, and En$\to$Es) under the conditions of the IWSLT Isometric Shared Task 2022. Using eight open-source large language models (LLMs) of varying sizes, we investigate how different prompting strategies, varying numbers of few-shot examples, and demonstration selection influence translation quality and length control. We discover that the phrasing of instructions, when aligned with the properties of the provided demonstrations, plays a crucial role in controlling the output length. Our experiments show that LLMs tend to produce shorter translations only when presented with extreme examples, while isometric demonstrations often lead to the models disregarding length constraints. While few-shot prompting generally enhances translation quality, further improvements are marginal across 5, 10, and 20-shot settings. Finally, considering multiple outputs allows to notably improve overall tradeoff between the length and quality, yielding state-of-the-art performance for some language pairs.

\end{abstract}

\section{Introduction}


Accurate and concise translations are increasingly needed in media applications such as subtitling \cite{matusov-etal-2019-customizing,karakanta-etal-2020-42} and dubbing \cite{federico-etal-2020-speech,lakew2021machine,tam22_interspeech,lakew2022isometric,rao-etal-2023-length}, where length constraints are critical. Dubbing, in particular, requires translations to stay within ±10\% of the source \change{character-level} length for seamless audio alignment \cite{lakew2022isometric}, a constraint known as \emph{isometric machine translation}. The 2022 Isometric MT Shared Task \cite{anastasopoulos-etal-2022-findings} found that most participating systems used lead tokens for length control, with some incorporating reranking or adjusted positional embeddings. Recent work also explored reinforcement learning for isometric English-Hindi MT \cite{mhaskar2024isometric} and examined length constraints in multiple language pairs \cite{bhavsar-etal-2022-hmist}.

Controlling translation length remains challenging compared to other constrained MT tasks, such as politeness \cite{sennrich-etal-2016-controlling} or diversity \cite{shu-etal-2019-generating} control. Previous approaches in encoder-decoder MT used length tokens \cite{lakew-etal-2019-controlling}, positional embeddings \cite{takase-okazaki-2019-positional,buet21_interspeech}, restricted search spaces \cite{niehues-2020-machine}, auxiliary length prediction tasks \cite{yang-etal-2020-predicting}, and explicit compression methods \cite{li2020explicit}.

\begin{figure}
    \centering
    \includegraphics[width=\linewidth]{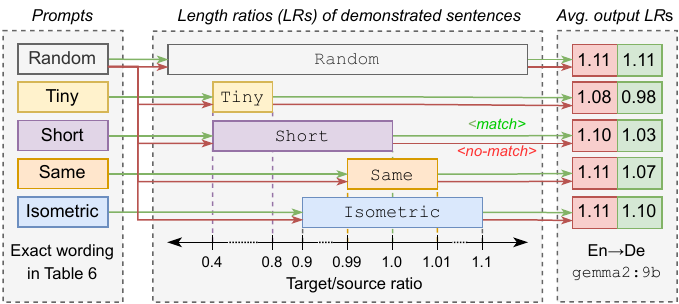}
    \caption{
    Overview of our experiment with prompts asking for different length constraints for the desired translation, complemented with few-shot examples demonstrating the given constraint (match) or not (no-match).
    Strong enough control to reach isometric translation needs matching instructions and preferably \emph{Tiny} or {Short} demonstrations.
    The construction of demonstration sets is described in \cref{sec:prompts} and the prompt content is presented in \cref{tab:prompt_templates} in \cref{app:prompts}.}
    \label{fig:match-vs-no-match-illustration}
\end{figure}

\begin{table*}[t]
    \centering
    \footnotesize
    \begin{tabular}{ll|ccc|ccc|ccc}
         & & \multicolumn{3}{c|}{En-De} & \multicolumn{3}{c|}{En-Fr} & \multicolumn{3}{c}{En-Es} \\\hline
         &Setup &LR&LC$\uparrow$& Count & LR&LC$\uparrow$& Count & LR&LC$\uparrow$& Count \\\hline\hline
Dev & Both  & 1.14$\pm$0.3 & 38.2 & 1415& 1.14$\pm$0.3 & 36.4 & 1412& 1.08$\pm$0.3 & 50.5 & 1316 \\\hline
         \multirow{2}{2em}{Test} & Development  & 1.15$\pm$0.2 & 37.5 & 200& 1.16$\pm$0.2 & 34.0 & 200& 1.03$\pm$0.2 & 58.0 & 200 \\
& Final Evaluation  & 1.03$\pm$0.2 & 65.5 & 200& 1.09$\pm$0.5 & 72.5 & 200& 0.98$\pm$0.2 & 64.0 & 200 \\\hline
    \end{tabular}
    \caption{The average target-to-source sample length ratio and its standard deviation (LR), length compliance (LC), i.e. the percentage of target-side sentences within a ±10\% range of the source character count, and the number of samples for two setups (\emph{Development} and \emph{Final Evaluation}) and for the testset (the MuST-C tst-COMMON and blind test sets) and the devset (MuST-C). The devset is used for selecting examples for few-shot prompting.}
    \label{tab:dataset_stats}
\end{table*}

\begin{table}[t]
    \centering
    \footnotesize
    \begin{tabular}{l|cccc}
         \multicolumn{5}{c}{En-De} \\\hline
         Pool type & Count & Min & Max & avg$\pm$std \\\hline\hline
\texttt{Random} & 1415 & 0.43 & 5.80 & 1.14$\pm$0.27 \\
\texttt{Isometric} & 537 & 0.90 & 1.10 & 1.02$\pm$0.05 \\
\texttt{Same} & 50 & 0.99 & 1.00 & 1.00$\pm$0.00 \\
\texttt{Short} & 343 & 0.43 & 1.00 & 0.90$\pm$0.11 \\
\texttt{Tiny} & 50 & 0.43 & 0.81 & 0.68$\pm$0.11 \\
    \end{tabular}
    \caption{Statistics of pools for En$\to$De: The number of samples, minimum and maximum target/source length ratio, and its average and standard deviation.}
    \label{tab:pool_stats_de}
\end{table}

With the rise of large language models (LLMs) \cite{radford2019language}, there has been a shift toward prompting \cite{vilar-etal-2023-prompting,zhang2023prompting,bawden-yvon-2023-investigating} and fine-tuning \cite{zhang-etal-2023-machine,moslem-etal-2023-adaptive} for MT. Prompting strategies notably affect performance, especially in few-shot settings \cite{vilar-etal-2023-prompting}. Studies found that randomly selected examples often improve results \cite{zhang2023prompting,bawden-yvon-2023-investigating}, though performance gains plateau beyond five examples \cite{chowdhery2023palm,vilar-etal-2023-prompting}. While models like \textsc{Bloom} tend to overgenerate in zero-shot settings \cite{bawden-yvon-2023-investigating}, fine-tuning methods such as QLoRA \cite{zhang-etal-2023-machine} have shown superior performance over few-shot learning. Real-time adaptive MT has also demonstrated strong results, with models like ChatGPT rivaling traditional MT systems \cite{moslem-etal-2023-adaptive,hendy2023good}. The use of LLMs for MT has led to the exploration of various prompt templates, with simple structures like `\texttt{[src]: [input] \textbackslash n [tgt]:}' proving effective \cite{zhang2023prompting,briakou-etal-2023-searching,zengglm}. The impact of example selection has also been examined, confirming that beyond five-shot settings, improvements become marginal \cite{garcia2023unreasonable,zhang2023prompting,chowdhery2023palm,vilar-etal-2023-prompting}.

Given these insights, we explore the application of LLMs to isometric MT, focusing on length control strategies. We analyze four prompting approaches: (1) uncontrolled translation, (2) isometric translation (±10\% length variation), (3) same-length translation, and (4) shorter translation, each paired with corresponding demonstration sets. Experiments are conducted on eight open-weight models (Llama~3, Gemma~2, Qwen~2 of two sizes each, and Mistral and Mixtral) across 0, 5, 10, and 20-shot settings for En$\to$De, En$\to$Fr, and En$\to$Es, following the 2022 Isometric Shared Task setup \citep{anastasopoulos-etal-2022-findings}.

Our results show that few-shot demonstrations affect translation outputs, but precise length control requires well-aligned instructions reflecting example properties, as summarized in \cref{fig:match-vs-no-match-illustration}. Additionally, we show that generating multiple outputs with different example sets substantially improves length control, matching competitive isometric MT systems and offering high potential for synthetic data creation in training encoder-decoder models. We publicly release all collected data for potential future analyses.\footnote{\url{https://github.com/J4VORSKY/Isometric-MT}}

\section{Experimental Setup \label{sec:experiments}}

\paragraph{Development} First, we conduct experiments with multiple settings (varying prompt type, the type of pools of demonstrations, and shot count in few-shot learning) to identify the best-performing configuration for length control. We refer to this as the \emph{Development} setup and use the following data:
\begin{itemize}
    \item \emph{Demonstration set}: We use the MuST-C dev-set for selecting few-shot examples. We choose the devset over the trainset to reserve the latter for potential future fine-tuning.
    \item \emph{Testset}: We use the first 200 examples from the MuST-C tst-COMMON, matching the number of examples in the evaluation blindset of the 2022 Isometric Shared Task.
\end{itemize}


\paragraph{Final Evaluation} We then use the best-performing setting from the \emph{Development} and evaluate it on the Isometric Shared Task test set:
\begin{itemize}
    \item \emph{Demonstration set}: We use the same demontration set as in the \emph{Development} setup.
    \item \emph{Testset}: We use the blindset from the IWSLT 2022 Isometric Shared Task, which consists of dialogues extracted from YouTube videos, totaling 200 examples.\footnote{\url{https://github.com/amazon-research/isometric-slt/tree/main/dataset}}
\end{itemize}


The statistics of the datasets used in both steps are displayed in \cref{tab:dataset_stats}.

\paragraph{Metrics}





Following the Isometric Shared Task, we use \emph{BERTScore}\footnote{\url{https://pypi.org/project/bert-score/0.3.11/}} \cite{zhang2019bertscore} to evaluate translation quality.
For completeness, we also report \emph{BLEU} \cite{papineni-etal-2002-bleu} scores  using sacreBLEU \cite{post-2018-call}.\footnote{\url{https://github.com/mjpost/sacrebleu}}$^,$\footnote{Signature: \texttt{nrefs:1|case:mixed|eff:no|tok:13a|} \texttt{smooth:exp|version:2.4.2}} We assess adherence to the ±10\% length constraint using the \emph{Length Compliance} (LC) metric \cite{anastasopoulos-etal-2022-findings}. Additionally, we report the average target-to-source \emph{Length Ratio} in \emph{Development} experiments and use it alongside Length Compliance in the \emph{Final Evaluation} to gauge length control.


\paragraph{Models}


We use the Ollama library\footnote{\url{https://ollama.com/library}} to load all models, which are provided in quantized versions (4-bit) without instruction fine-tuning (more details in \cref{app:generation}). Models used in our experiments include: \texttt{llama3:8b}, \texttt{llama3:70b} \cite{dubey2024llama}; \texttt{gemma2:9b}, \texttt{gemma2:27b} \cite{team2024gemma}; \texttt{qwen2:7b}, \texttt{qwen2:72b} \cite{yang2024qwen2}; \texttt{mistral:7b} \cite{jiang2023mistral} and \texttt{mixtral:8x7b} \cite{jiang2024mixtral}. For detailed descriptions, refer to the original papers.

\section{Prompts}
\label{sec:prompts}




In our experiments, we use English as the language of the prompts \cite{zhang2023prompting} and explicitly specify the source and target languages within the prompt \cite{zhang2023prompting, bawden-yvon-2023-investigating}. Our focus is on length control when testing various prompt formulations. While large language models (LLMs) show strong performance in machine translation, they sometimes lag behind supervised neural models \cite{zhang2023prompting, chowdhery2023palm, kocmi-etal-2023-findings}. To our knowledge, length control has not been extensively explored for LLMs in machine translation.

\paragraph{Prompt construction} We construct prompts by concatenating template parts and replacing placeholders with the appropriate values. The \emph{Random} (uncontrolled) template instructs the model to generate a translation of the source sentence without any length restrictions. In the \emph{Isometric} template, the model is instructed to generate a translation within ±10\% of the source text's character count. The \emph{Same} template instructs the model to produce a translation that exactly matches the source text length, while the \emph{Short / Tiny} template directs the model to generate a shorter translation, as the length ratios between studied language pairs often exceed 1, and a standard translation (typically longer) is not desired. A detailed overview of the prompt templates is in \cref{tab:prompt_templates} in \cref{app:prompts}.

\begin{table*}[t]
    \centering
    \footnotesize
    \begin{tabular}{l|cc||cc||cc||cc||cc}
        \multicolumn{11}{c}{En-De} \\\hline
        & \multicolumn{2}{c||}{\texttt{Random}} & \multicolumn{2}{c||}{\texttt{Isometric}} & \multicolumn{2}{c||}{\texttt{Same}} & \multicolumn{2}{c||}{\texttt{Short}} & \multicolumn{2}{c}{\texttt{Tiny}} \\\hline\hline
        Model $\backslash$ Match & No & Yes & No & Yes &  No & Yes & No & Yes & No & Yes \\\hline
        \texttt{gemma2:27b} & 1.100 & 1.097 & 1.099 & 1.094 & 1.098 & 1.097 & \underline{1.087} & \underline{1.011} & \underline{1.066} & \underline{0.955} \\
\texttt{gemma2:9b} & 1.108 & 1.106 & 1.106 & 1.101 & \underline{1.106} & \underline{1.073} & \underline{1.099} & \underline{1.026} & \underline{1.080} & \underline{0.981} \\
\texttt{llama3:70b} & 1.149 & 1.151 & 1.149 & 1.139 & 1.141 & 1.134 & \underline{1.138} & \underline{1.005} & \underline{1.122} & \underline{0.905} \\
\texttt{llama3:8b} & 1.106 & 1.100 & 1.093 & 1.108 & 1.099 & 1.112 & \underline{1.085} & \underline{1.048} & \underline{1.056} & \underline{0.994} \\
\texttt{mistral:7b} & 1.133 & 1.129 & 1.126 & 1.128 & 1.135 & 1.125 & 1.121 & 1.105 & \underline{1.138} & \underline{1.085} \\
\texttt{mixtral:8x7b} & 1.402 & 1.411 & 1.375 & 1.362 & 1.378 & 1.381 & 1.385 & 1.297 & 1.363 & 1.265 \\
\texttt{qwen2:72b} & 1.223 & 1.169 & 1.195 & 1.178 & 1.184 & 1.173 & 1.170 & 1.128 & 1.164 & 1.129 \\
\texttt{qwen2:7b} & 1.132 & 1.160 & 1.144 & 1.125 & 1.129 & 1.129 & 1.128 & 1.135 & 1.117 & 1.095 \\
    \end{tabular}
    \caption{The evaluation is conducted as follows: We first compute the average target length per input sentence across 10 runs. Next, we calculate the target-to-source length ratio for each instance and average these values for each pool type. The results are reported separately for cases where the instructions match (`Yes') or do not match (`No') the sample properties in 5-shot prompting. Differences with a $p$-value < 0.1 for each pool type are underlined.}
    \label{tab:difference_match_vs_no_match_de}
\end{table*}

We evaluate models in zero-shot and few-shot settings. They often overgenerate, adding explanations or extra translations, as noted by \citet{bawden-yvon-2023-investigating}. While these authors used regular expressions to extract translations, we prevent this by explicitly instructing models to output only the translation, which proves effective. Further analysis is in \cref{app:overgeneration}.

\paragraph{Sample Selection} In preparing examples for the few-shot setting, we construct sampling pools by filtering the demonstration set based on the following criteria: \texttt{Random} selects samples without any filtering; \texttt{Isometric} contains only examples with a target-to-source length ratio within ±10\%; \texttt{Same} sorts references by increasing $|r-1.0|$ (where $r$ is the length ratio) and selects the top $N=50$ instances; \texttt{Short} selects samples with target-to-source ratios in the range $[0, 1]$; \texttt{Tiny} samples the 50 examples with the smallest target-to-source ratio. The illustration is in \cref{fig:match-vs-no-match-illustration}.

Statistics for each sampling pool for En$\to$De are in~\cref{tab:pool_stats_de}. As other languages follow the same trend, their statistics are in \cref{tab:pool_stats} in \cref{app:pool_stats}. Following \citet{zhang2023prompting}, we use the following template for in-context samples: \texttt{[src lang]}: \texttt{[src sentence]} $\neg$ \texttt{[tgt lang]}: \texttt{[tgt sentence]}. 

\begin{figure}[t]
    \centering
    \includegraphics[width=\linewidth]{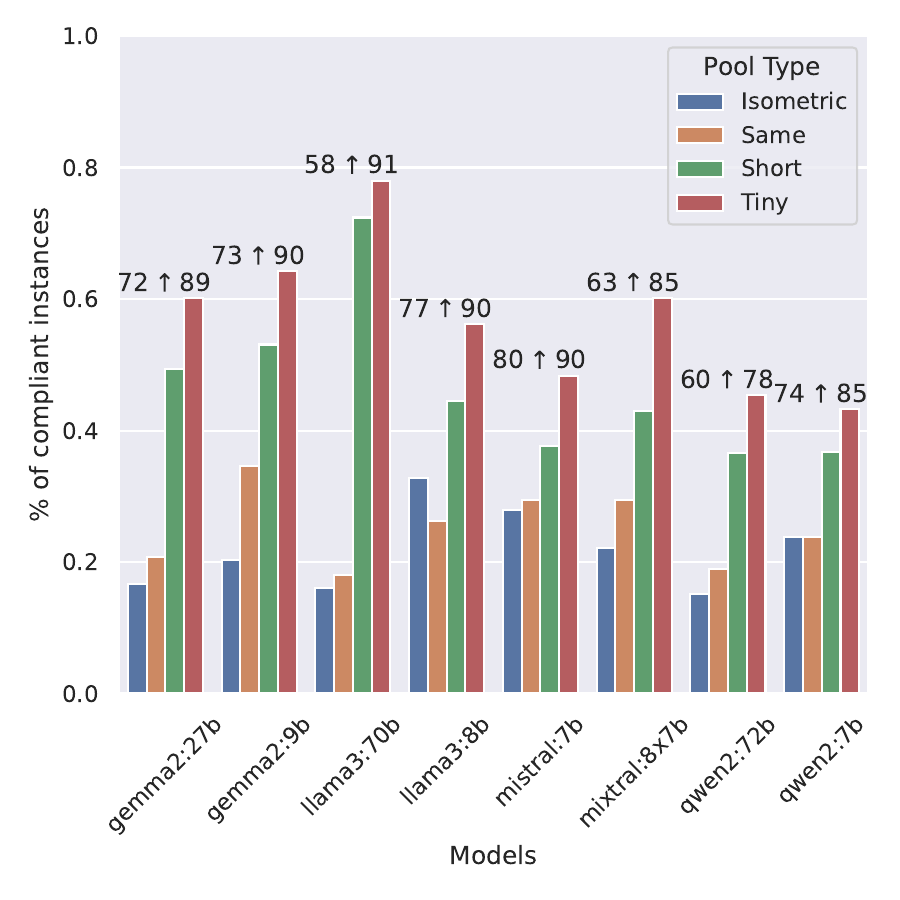}
    \caption{The percentage of input sentences (across all language directions) for which at least one of ten generated translations meets the isometric condition when the model is prompted to produce isometric, same-length, short, and tiny outputs aligned with respective 5-shot demonstration sets. This evaluation is restricted to input sentences where the particular model did not generate any isometric translation in ten attempts using the uncontrolled prompt.
    }
    \label{fig:comliance_proportion}
\end{figure}

\section{Analysis \label{sec:analysis}}

In all experiments, the prompts remain identical across all models within a given setting. To reduce the bias of sampling from demonstration sets, we performed 10 runs for every setting.

\subsection{Prompt and Pool Type Relation}

First, we analyze how much the selection of examples is related to the instruction provided in the prompt in the few-shot prompting and how this combination influences the translation length. We therefore compare two setups:

\paragraph{Prompt and Pool Type Match} We create matching pairs of prompts and pool types as follows: Random--\texttt{Random}, Isometric--\texttt{Isometric}, Same--\texttt{Same}, Short--\texttt{Short}, Tiny--\texttt{Tiny}.

\paragraph{Prompt and Pool Type Mismatch} We keep the Random prompt for all pool types.

\begin{figure*}[t!]
    \centering
    \includegraphics[width=\linewidth]{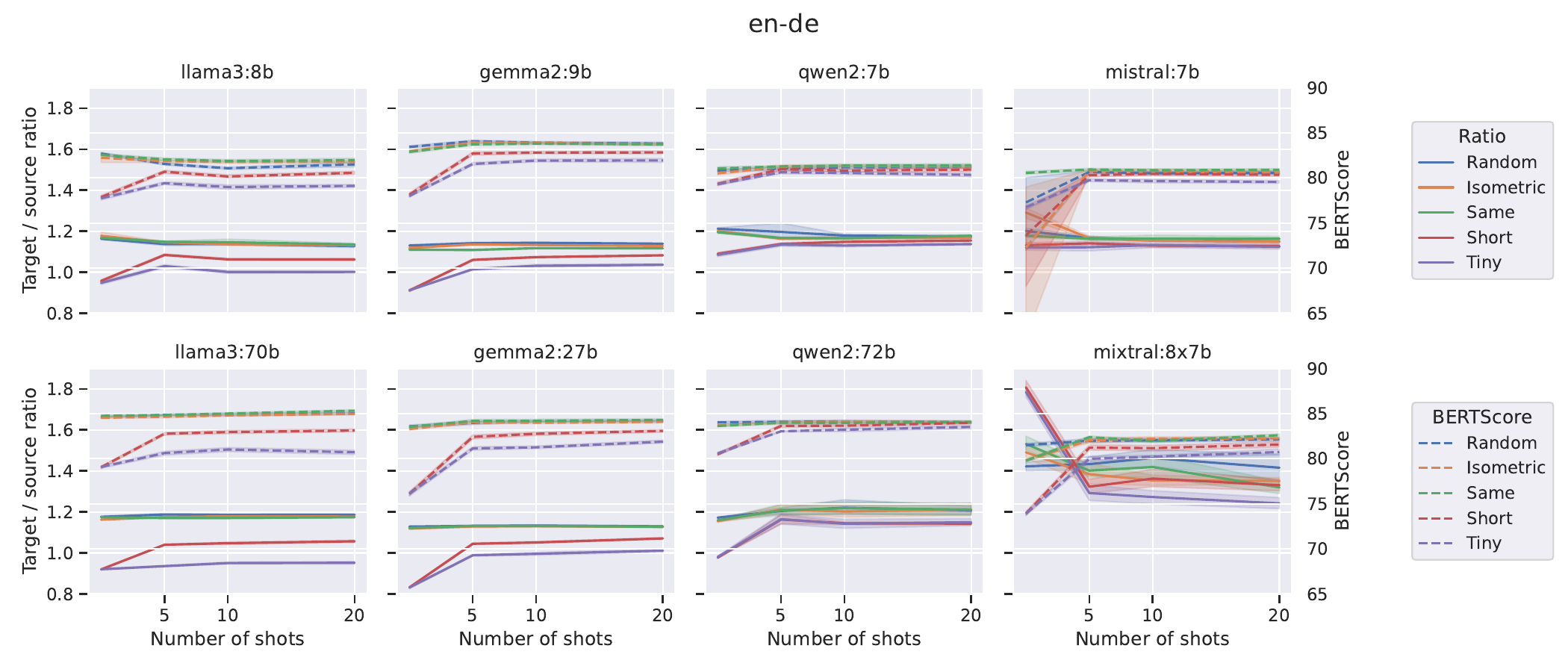}
    \caption{En$\to$De translation quality (BERTScore, dashed lines and the right hand y-axes) and length ratio (solid lines and left-hand y-axes) for all few-shot settings, models and language pairs.}
    \label{fig:bs_ratio_pools_de}
\end{figure*}

\begin{figure*}[t!]
    \centering
    \includegraphics[width=\linewidth]{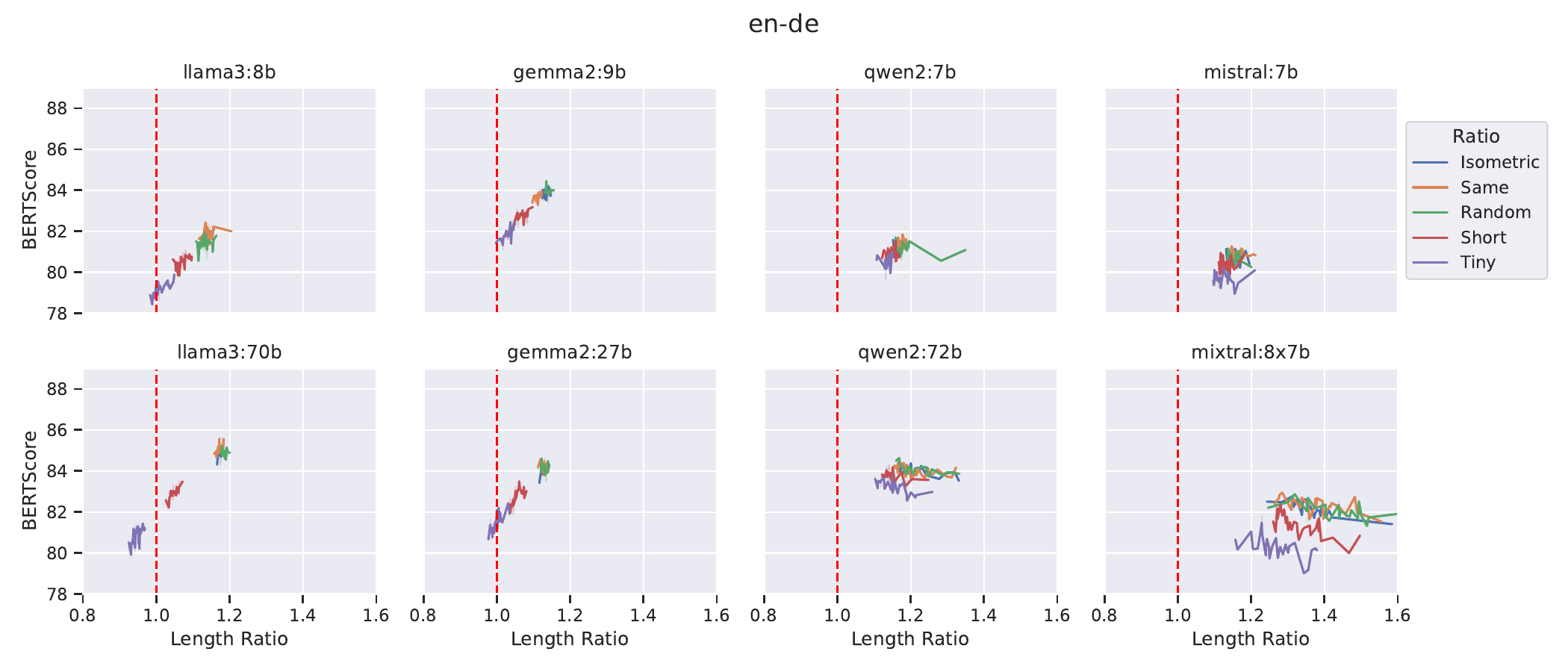}
    \caption{En$\to$De trade-off between the length ratio (x-axis) and translation quality (y-axis) for 5, 10, 20-shot settings and all models.}
    \label{fig:tradeoff_de}
\end{figure*}

We compare these two configurations for En$\to$De in \cref{tab:difference_match_vs_no_match_de}, the remaining translation directions are documented in \cref{app:match-vs-non-match}. Our results indicate that the length ratios are mostly affected when the instruction aligns with the pool type, compared to when there is no such match (we can also see a tendency to generate shorter outputs when comparing ``no alignment'' columns across different pool types, but the difference is negligible). This match-versus-no-match difference is statistically significant in Gemma and Llama models, particularly for the \texttt{Short} and \texttt{Tiny} pools. Additionally, the \texttt{Isometric} and \texttt{Same} pools do not appear to induce shorter translations compared to random sampling, as evidenced by the similar values observed in the first three columns. We hypothesize that requesting outputs to preserve the input length somehow guides models to reproduce the distribution of the training data rather than actually considering the length (i.e. models implicitly assume that typical translation is of the same length). However, in studied language directions, what is considered as normal ratio, is skewed towards values greater than 1. In other words, models naturally follow the length distribution they were trained on and can overcome this bias only when extreme examples are provided.

To further highlight the utility of our approach, \cref{fig:comliance_proportion} focuses on cases where models consistently fail to produce isometric translations under the \emph{Random}-\texttt{Random} setting, even after 10 runs. This occurs in about 30\% of devset sentences on average. The figure shows how alternative prompts improve length compliance, with \emph{Tiny} and \emph{Short} settings achieving up to 80\% isometric translations for \texttt{Llama3:70b} when at least one of 10 runs succeeds.
The overall practical ability of each of the models to achieve isometric translation is summarized by the two numbers above the bars in \cref{fig:comliance_proportion}. The first number indicates the percentage of devset sentences that were translated in a compliant way by default and the second number indicates to which proportion we raised this using the \emph{Tiny} prompt. Note that in the worst case, this level of compliance is reached at 20x the translation cost (10 attempts by default plus 10 \emph{Tiny} attempts). In practice, however, we can switch to the \emph{Tiny} prompt after the first unsuccessful attempt in the default generation. The number of additional generations with \emph{Tiny} setting depends on resource constraints and requirements. But even after just one attempt, \texttt{Llama3:70b} achieves isometric translations in 35\% of cases. Full results are in \cref{app:match-vs-non-match}.


\subsection{Comparing Demonstration Pools}

\begin{table*}[t]
    \centering
    \footnotesize
    \setlength{\tabcolsep}{4pt}
    \begin{tabular}{l|cccc|cccc|cccc}
& \multicolumn{4}{c|}{En$\to$De} & \multicolumn{4}{c|}{En$\to$Fr} & \multicolumn{4}{c}{En$\to$Es} \\\hline
System & LR$\downarrow$ & LC$\uparrow$ & BS$\uparrow$ & BLEU$\uparrow$ & LR$\downarrow$ & LC$\uparrow$ & BS$\uparrow$ & BLEU$\uparrow$ & LR$\downarrow$ & LC$\uparrow$ & BS$\uparrow$ & BLEU$\uparrow$ \\
\hline\hline
    \textsc{StrongBaseline} & \bf 1.03 & 68.0 & 77.44 & 21.6 & 1.02 & 75.5 & \bf 81.75& \bf 36.2& \bf 1.00 & 80.5 & 81.86 & 36 \\
    \textsc{AppTek}-Constrained & 1.11 & 86.5 & 77.32 & 18.7 & - & - & - &- &- &- &- &- \\
    \textsc{NUV}-Unconstrained & - & - & - & - & 1.10 & 47.5 & 79.96 & 27.1 &- &- &- &- \\
    \textsc{HW-TSC}-Unconstrained & \bf 1.03 & 96.5 & 75.79 & 20.2 & - & - & - &- &- &- &- &- \\
    \textsc{HW-TSC}-Constrained & 1.28 & \bf 98.0 & 74.07 & 17.9 & 1.19 & \bf 96.0 &  76.11& 31.5 & 1.18 & \bf 96.5 & 78.57 & 29.9 \\
    \textsc{APV}-Unconstrained & 1.68 & 39.0 & 73.68 & 16.5 & 1.21 & 45.0  & 77.77 & 32.9 & 1.05 & 49.5 & 80.87 & 35.3 \\
    \textsc{WeakBaseline} & 1.29 & 43.0 & 74.86 & 15.5 & 1.48 & 37.0 & 77.18& 25.2 & 1.38 & 51.0 & 78.32 & 27.7 \\\hline\hline
    \texttt{model=gemma2:27b-k=1} & 1.07 & 43.5 & 77.08 & 19.0 & 1.05 & 47.5 & 78.30 & 32.7 & \bf 1.00 & 55.5 & 83.20 & 40.3 \\
\texttt{model=gemma2:27b-k=3} & 1.08 & 58.0 & 77.96 & 20.2 & 1.07 & 60.5 & 79.96 & 33.5 & 1.01 & 66.5 & 83.29 & 40.3 \\
\texttt{model=gemma2:27b-k=5} & 1.09 & 62.5 & \bf 77.98 & 20.4 & 1.06 & 62.5 & 80.01 & 33.9 & 1.02 & 68.0 & 83.16 & 40.0 \\
\texttt{model=gemma2:27b-k=10} & 1.08 & 68.5 & 77.84 & \bf 21.9 & 1.08 & 69.0 & 80.05 & 35.6 & 1.01 & 70.5 & \bf 83.62 & \bf 40.8 \\\hline
\texttt{model=gemma2:9b-k=1} & 2.24 & 42.5 & 77.04 & 17.7 & 0.00 & 0.0 & 0.00 & 0.0 & 1.02 & 54.5 & 82.47 & 39.0 \\
\texttt{model=gemma2:9b-k=3} & 1.19 & 58.5 & 77.24 & 20.6 & 1.07 & 60.5 & 80.38 & 34.1 & 1.03 & 65.5 & 83.41 & 36.8 \\
\texttt{model=gemma2:9b-k=5} & 1.07 & 64.5 & 77.38 & 20.9 & 1.06 & 65.5 & 80.66 & 35.5 & 1.03 & 73.0 & 83.30 & 37.2 \\
\texttt{model=gemma2:9b-k=10} & 1.08 & 64.0 & 77.48 & 21.7 & 1.06 & 70.5 & 80.72 & 34.9 & 1.03 & 73.0 & 83.17 & 37.6 \\\hline
\texttt{model=llama3:70b-k=1} & 1.09 & 49.0 & 76.57 & 20.9 & 1.05 & 41.0 & 76.44 & 28.6 & 0.96 & 46.5 & 79.29 & 31.4 \\
\texttt{model=llama3:70b-k=3} & 1.06 & 62.5 & 77.18 & 22.1 & \bf 1.00 & 55.5 & 77.64 & 30.9 & 1.03 & 59.5 & 80.64 & 34.4 \\
\texttt{model=llama3:70b-k=5} & 1.06 & 65.0 & 77.24 & 22.2 & 1.02 & 64.0 & 77.62 & 32.5 & 1.02 & 65.5 & 80.96 & 35.1 \\
\texttt{model=llama3:70b-k=10} & 1.07 & 69.0 & 77.23 & 21.7 & 1.04 & 68.0 & 78.24 & 33.7 & 1.02 & 70.5 & 81.37 & 35.8 \\\hline\hline
\texttt{model=llama3:8b-k=1} & 1.21 & 42.0 & 74.30 & 13.8 & 1.16 & 47.5 & 74.71 & 22.9 & 1.03 & 48.5 & 77.80 & 28.6 \\
\texttt{model=llama3:8b-k=3} & 1.09 & 56.0 & 75.79 & 15.9 & 1.09 & 60.5 & 75.96 & 25.5 & 0.99 & 65.0 & 79.76 & 30.6 \\
\texttt{model=llama3:8b-k=5} & 1.09 & 60.5 & 76.10 & 16.7 & 1.10 & 69.5 & 76.32 & 26.1 & 1.01 & 69.5 & 80.15 & 31.4 \\
\texttt{model=llama3:8b-k=10} & 1.09 & 65.0 & 76.28 & 16.9 & 1.08 & 75.0 & 77.02 & 26.2 & 1.03 & 74.0 & 79.88 & 29.0 \\\hline\hline
    OracleBLEU & 1.04 & 78.0 & 80.82 & 37.6 & 1.03 & 85.0 & 83.93 & 52.9 & 1.01 & 88.5 & 87.01 & 57.5 \\
    \end{tabular}
    \caption{\emph{Final Evaluation} --- Length Ratio (LR), Length Compliance (LC), BERTScore (BS) and BLEU --- of the best setting (10-shot, pool type \texttt{Tiny}) across different Llama and Gemma models compared to the submissions of IWSLT Isometric Shared Task. The $k$ values indicate the number of demonstration sampling runs (i.e. different outputs) from which we select the best one using COMETKIWI. To avoid any possible evaluation difference, we (re-)evaluated all the outputs, ours and IWSLT22 ones, using the script provided by the organizers of the shared task. The best results are in bold.}
    \label{tab:final_evaluation}
\end{table*}

We give below a more detailed evaluation across all few-shot settings, models, and pool types, using only settings when the instruction matches the pool type and where the model is instructed to output only the translation. Both length ratios and BERTScore values are reported for En$\to$De, with the results presented in \cref{fig:bs_ratio_pools_de}. For a comprehensive view of all few-shot settings, detailed numerical results are reported in \cref{app:few-shot}.

\paragraph{Length Ratios and Length Control}
In terms of length ratio, all models consistently exhibit the same trend: the ratios are highest for random sampling, followed by isometric sampling, and then by shorter examples. Providing extreme examples encourages models to produce shorter translations. Interestingly, in the zero-shot setting, we observe a length ratio lower than 1.0 for the Llama and Gemma models.
However, when demonstrations are also given in few-shot settings for these models, translations are longer, even when the associated demonstrations are short or very short.

\paragraph{Few-shot Prompting}
Another notable observation is that increasing the number of examples in few-shot prompting does not substantially enhance regular translation quality (i.e., translation without length restrictions), which is consistent with previous findings \cite{bawden-yvon-2023-investigating, zhang2023prompting, chowdhery2023palm}. Including shorter examples sometimes improves adherence to length limitations (e.g., for \texttt{llama3:8b}); this effect is not observed for all models (e.g., for \texttt{gemma2:27b}).

\paragraph{Translation Quality Scores}
The largest translation quality scores are observed when the unfiltered pool (\texttt{Random}) is used, which is expected as this corresponds to an unconstrained setting. The top-performing model in terms of BERTScore for English-German translations is \texttt{llama3:70b}. For the other language pairs, \texttt{gemma2:9b}, \texttt{gemma2:27b} and \texttt{qwen2:72b} achieve the largest translation score.

\paragraph{Length Ratio and Translation Quality Tradeoff} We also compare translation scores with length ratios. The results are presented in \cref{fig:tradeoff_de} for En$\to$De direction (the rest in \cref{fig:tradeoff} in \cref{app:tradeoff}). We can see that only Llama and Gemma models are capable of reaching 1.0 length ratio. Our results also highlight the impact of model size on performance, with larger models consistently outperforming their smaller counterparts in BERTScore, except for \texttt{gemma2:9b} which reports similar performance to its large counterpart.

\section{Final Evaluation}

Since Llama and Gemma models achieve the best performance on all language pairs on average, we select and evaluate them in \emph{Final Evaluation} using the Isometric Shared Task blind set. We generate outputs using 10 distinct sets of 10 examples (10-shot), each drawn from the \texttt{Tiny} pool as it yields the best length control. We keep only outputs in the $\pm$10\% length constraint\footnote{If none of the translations adhere to the $\pm$10\% length constraint, we keep the original unfiltered set.} and select the best one according to reference-free COMET, i.e. COMETKIWI\footnote{\url{https://huggingface.co/Unbabel/wmt22-cometkiwi-da}} score \cite{rei-etal-2022-cometkiwi}. We then compare these results with the submissions from the IWSLT 2022 Isometric Shared Task, specifically those from the APPTEK \cite{wilken-matusov-2022-appteks}, HW-TSC \cite{li-etal-2022-hw}, Amazon Prime Video (APV), and NUV teams \cite{bhatnagar-etal-2022-hierarchical}, in addition to the two (strong and weak) baselines provided by the organizers. For a brief overview of each system, please refer to \citet{anastasopoulos-etal-2022-findings}. Additionally, we compare our results to an OracleBLEU setting, where the best translation is selected according to the sentence BLEU score across all configurations after filtering out translations that fall outside ±10\% of the source character count. The results are summarized in \cref{tab:final_evaluation}.

Our results show that for the En$\to$De and En$\to$Es language pairs, the Gemma models achieve output quality comparable to the strong baseline. While the translation quality metrics surpass that of the strong baseline, the length control is slightly less precise. For En$\to$Fr, however, the strong baseline continues to outperform our models in terms of quality as well as LR. Although generating 10 different outputs for each source sentence may not be feasible in practice, this approach could be beneficial for producing synthetic data for training isometric machine translation models.


\section{Conclusion}

In this paper, we explored the use of LLMs for isometric machine translation, focusing on strategies to control the translation length.
Our key findings are as follows: First, effective length control in few-shot prompting requires the simultaneous use of appropriate demonstrations and matching instructions.
Second, generating multiple outputs achieves the best trade-off between length control and translation quality, indicating the high capability of LLMs to generate desired outputs. It might be also useful for creating synthetic training data. Although prompting 10 times may seem inefficient, it would not be necessary for every sample in practice. Since half of the samples are already length-compliant --- even with the uncontrolled \emph{Random} prompt --- compliance for the rest can be achieved iteratively by generating translations until the length constraint is met.
Future work might benefit from fine-tuning LLMs or from a more in-depth analysis of the internal representation of length in LLMs to avoid many samples to generate.







\section*{Limitations}

We compare our results primarily with system submissions from the Isometric Shared Task 2022, as more recent models either do not address the language pairs examined in this study (e.g., Hindi-English by \citet{mhaskar2024isometric}) or are not publicly available \cite{bhavsar-etal-2022-hmist}. Additionally, we do not evaluate performance on any downstream tasks, such as subtitling or dubbing.

We did not conduct a detailed analysis of ensemble methods, particularly concerning ensembling across different models or pool types. Moreover, for the \texttt{Tiny} and \texttt{Same} pools, we do not analyze the effect of varying $N$.

When collecting 10 outputs in \emph{Final Evaluation}, the associated computational cost increases considerably. While this approach may not be feasible for real-world applications, it can be valuable for generating high-quality examples for isometric machine translation model training. To further reduce computational costs, one could regenerate only those translations that do not meet the specified length constraints.

Finally, it is important to note that we exclusively used quantized versions of the models in our experiments, likely resulting in sub-optimal translation scores.


\section*{Acknowledgements}

\change{The work has been partially supported by the grant 272323 of the Grant Agency of Charles University, SVV project number 260 821 and by the grant CZ.02.01.01/00/23\_020/0008518 (``Jazykověda, umělá inteligence a jazykové a řečové technologie: od výzkumu k aplikacím'').}

\bibliography{anthology,custom}

\appendix

\section{Overgeneration}
\label{app:overgeneration}

\citet{bawden-yvon-2023-investigating} have demonstrated that the \textsc{Bloom} model tends to overgenerate, specifically it continues to produce translations in additional languages beyond the desired output. In our preliminary experiments, we observed similar behavior across several models, which manifested in two distinct ways: (1) models frequently provided explanations alongside the translation, and (2) models embedded the translation within a broader text.

\begin{figure}[t]
    \centering
    \includegraphics[width=0.9\linewidth]{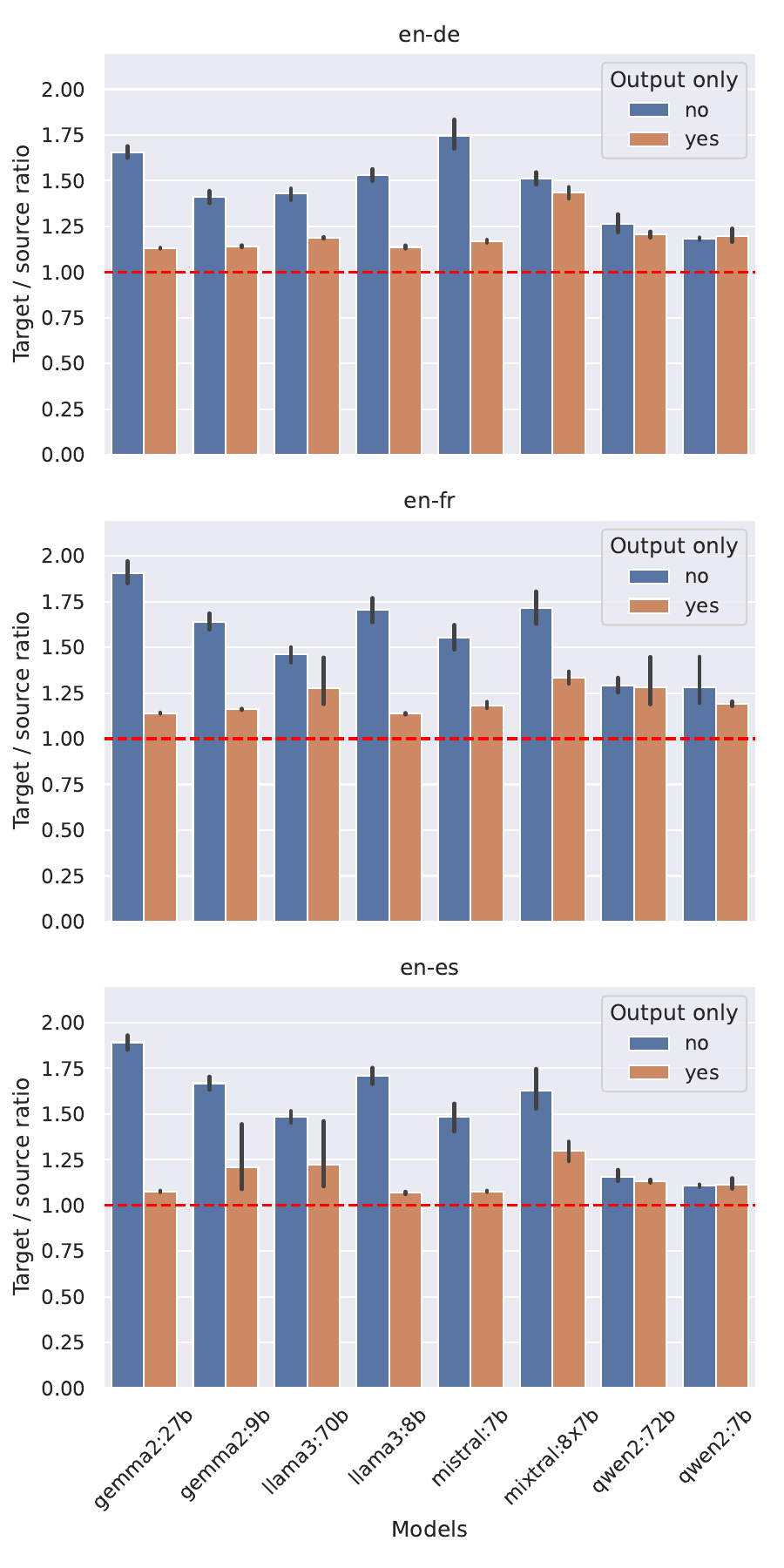}
    \caption{Restricted vs unrestricted prompt for 5-shot examples and the random pool when we discard everything after the first new line. In restricted, we add `output translation only' at the end of the prompt. The red dashed line corresponds to a ratio of 1.0.}
    \label{fig:restricted_vs_unrestricted_prompt_truncated}
\end{figure}

To mitigate the issue of overgeneration, we implemented a straightforward yet highly effective solution. Specifically, we appended an instruction to the prompt, explicitly directing the model to output only the translation. This approach proved to be remarkably effective, obviating the need for more complex techniques such as truncation or the application of regular expressions to filter the translation. We evaluated the impact of this method on translation length within a 5-shot setting, utilizing a randomly selected pool with uncontrolled instruction types. For each model, we constructed 10 distinct prompts with different examples, and we discarded generated text after the first new line character because at this place an explanation often begins. The averaged length ratios are presented in \cref{fig:restricted_vs_unrestricted_prompt_truncated}.

\begin{table}[t]
    \centering
    \footnotesize
    \begin{tabular}{l|cc|cc|cc}
         & \multicolumn{2}{c|}{En-De} & \multicolumn{2}{c|}{En-Fr} & \multicolumn{2}{c}{En-Es}\\\hline\hline
         Model & No & Yes &  No & Yes & No & Yes \\\hline
\texttt{gemma2:9b} & 100 & \bf 0 & 100 & \bf 2 & 100 & \bf 2 \\
\texttt{gemma2:27b} & 100 & \bf 1 & 100 & \bf 2 & 100 & \bf 1 \\
\texttt{llama3:8b} & 100 & \bf 3 & 100 & \bf 3 & 100 & \bf 1 \\
\texttt{llama3:70b} & 100 & \bf 0 & 99 & \bf 0 & 99 & \bf 0 \\
\texttt{qwen2:7b} & \bf 0 & \bf 0 & \bf 0 & \bf 0 & \bf 0 & \bf 0 \\
\texttt{qwen2:72b} & 12 & \bf 3 & 7 & \bf 1 & 5 & \bf 1 \\
\texttt{mistral:7b} & 16 & \bf 3 & 9 & \bf 1 & 19 & \bf 1 \\
\texttt{mixtral:8x7b} & 38 & \bf 15 & 29 & \bf 15 & 36 & \bf 20 \\
    \end{tabular}
    \caption{The average percentage of translations that contain a new line, indicating overgeneration (5-shot setting). `Yes' and `No' columns denote the restricted and unrestricted prompt, respectively.}
    \label{tab:restricted_vs_unrestricted_nl}
\end{table}

The results indicate that our approach maintains length consistency across all models and language pairs, with values remaining close to 1.0. The only exceptions are \texttt{mixtral:8x7b}, which tends to generate longer text even with the restrictive instruction, and \texttt{qwen2:7b}, which is the only one that does not tend to overgenerate in the first place.


Since overgeneration, when it does occur, typically manifests itself as additional text generated after a newline character, we counted the occurrences of text generation following a newline in both the restricted and unrestricted settings to further evaluate the effectiveness of our method. As shown in \cref{tab:restricted_vs_unrestricted_nl}, in many instances --- particularly with the Llama and Gemma models --- there is a clear tendency for models to generate explanatory text after a newline when output is unrestricted. Conversely, when the output is restricted to translation only, the occurrence of additional text is substantially reduced. Based on these observations, we adopted this restrictive instruction in all subsequent experiments and we also ignore any output after the newline character.

\paragraph{Examples}
\label{app:examples}
Examples of overgeneration where (1) models frequently provided explanations alongside the translation, and (2) models embedded the translation within a broader text:
\begin{enumerate}
    \item \texttt{qwen2:72b}: \textit{\dots{}English: Not surprisingly, this destruction also endangers bonobo survival. $\neg$ German:}  Überraschenderweise gefährdet dieser Niedergang auch das Überleben der Bonobos. $\neg$ $\neg$ However, a more accurate translation would be: $\neg$ Unüberraschenderweise gefährdet diese Zerstörung auch das Überleben der Bonobos.
    \item \texttt{llama3:8b}: \textit{\dots{}English: But still it was a real footrace against the other volunteers to get to the captain in charge to find out what our assignments would be. $\neg$ Spanish:} Based on the provided examples, here is a possible translation: $\neg$ $\neg$ Spanish: Pero todavía fue un verdadero carrera contra los otros voluntarios para llegar al capitán al mando y encontrar qué serían nuestras asignaciones.$\neg$ $\neg$ This translation takes into account the nuances of the original sentence\dots{} \textit{(explanation continues)}
\end{enumerate}




\section{Generation Details}
\label{app:generation}

\subsection{Inference Hyperparameters}

In all experiments, text generation uses multinomial sampling, with default parameters provided by the Ollama library: top-K 40 sampling ($K=40$) and a temperature of $0.8$. Generation stops after 512 tokens or when \texttt{<EOT>} (end of turn) token is printed.

\subsection{Prompt Templates}
\label{app:prompts}

\begin{table*}[t]
    \centering
    \footnotesize 
    \begin{tabular}{l|l|l}
        Part & Prompt type & Zero-shot\\\hline\hline
        1 & - & Translate the following text from \texttt{[src lang]} into \texttt{[tgt lang]} \\\hline
        \multirow{4}{2em}{2} & Random & .\\
        & Isometric &ensuring that it is within ±10\% of the character count of the source. \\
        & Same & ensuring that it has the same length as the source. \\
        & Short / Tiny & ensuring that it is shorter than the source. \\\hline
        \multirow{2}{2em}{3} & No & $\neg$ \\
        & Yes & Output only the translation. $\neg$ \\\hline
        4 & - & \texttt{[src lang]}: \texttt{[src sentence]} $\neg$ \texttt{[tgt lang]}:\\
        \multicolumn{3}{c}{}\\
        Part & Prompt type & Few-shot\\\hline\hline
        1 & - & Here are examples of translations in \texttt{[tgt lang]} \\\hline
        \multirow{4}{2em}{2} & Random &  of the source in \texttt{[src lang]}: $\neg$\\
        & Isometric & that are within ±10\% of the character count of the source in \texttt{[src lang]}: $\neg$\\& Same & that have the same length as the source in \texttt{[src lang]}: $\neg$\\
        & Short / Tiny & that are shorter than the source in \texttt{[src lang]}: $\neg$\\\hline
        3 & - & $N \times$ \{\texttt{[src lang]}: \texttt{[src sentence]} $\neg$ \texttt{[tgt lang]}: \texttt{[tgt sentence]} $\neg$\}\\\hline
        4 & - & Provide translation for the following sentence given the examples above. \\\hline
        \multirow{2}{2em}{5} & No & $\neg$ \\
        & Yes & Output only the translation. $\neg$ \\\hline
        6 & - & \texttt{[src lang]}: \texttt{[src sentence]} $\neg$ \texttt{[tgt lang]}:\\
    \end{tabular}
    \caption{Zero-shot (upper) and few-shot (lower) prompt templates. $\neg$ stands for new line. Actual prompts are constructed by sequentially concatenating prompt \emph{parts} (1--6).}
    \label{tab:prompt_templates}
\end{table*}

The construction of templates is depicted in \cref{tab:prompt_templates}. The prompts are created by concatenating prompt parts (1--6).

\subsection{Pool Statistics}
\label{app:pool_stats}

\begin{table}[t]
    \centering
    \footnotesize
    \begin{tabular}{l|cccc}
         \multicolumn{5}{c}{En-De} \\\hline
         Pool type & Count & Min & Max & avg$\pm$std \\\hline\hline
\texttt{Random} & 1415 & 0.43 & 5.80 & 1.14$\pm$0.27 \\
\texttt{Isometric} & 537 & 0.90 & 1.10 & 1.02$\pm$0.05 \\
\texttt{Same} & 50 & 0.99 & 1.00 & 1.00$\pm$0.00 \\
\texttt{Short} & 343 & 0.43 & 1.00 & 0.90$\pm$0.11 \\
\texttt{Tiny} & 50 & 0.43 & 0.81 & 0.68$\pm$0.11 \\
\multicolumn{5}{c}{}\\
\multicolumn{5}{c}{En-Fr} \\\hline
\texttt{Random} & 1412 & 0.29 & 4.90 & 1.14$\pm$0.28 \\
\texttt{Isometric} & 505 & 0.90 & 1.10 & 1.02$\pm$0.05 \\
\texttt{Same} & 50 & 0.99 & 1.00 & 1.00$\pm$0.00 \\
\texttt{Short} & 348 & 0.29 & 1.00 & 0.88$\pm$0.14 \\
\texttt{Tiny} & 50 & 0.29 & 0.76 & 0.60$\pm$0.13 \\
\multicolumn{5}{c}{}\\
\multicolumn{5}{c}{En-Es} \\\hline
\texttt{Random} & 1316 & 0.30 & 5.70 & 1.08$\pm$0.32 \\ 
\texttt{Isometric} & 659 & 0.90 & 1.10 & 1.01$\pm$0.05 \\
\texttt{Same} & 50 & 1.00 & 1.00 & 1.00$\pm$0.00 \\
\texttt{Short} & 490 & 0.30 & 1.00 & 0.89$\pm$0.12 \\
\texttt{Tiny} & 50 & 0.30 & 0.72 & 0.59$\pm$0.12 \\
    \end{tabular}
    \caption{Statistics of pools: The number of samples, minimum and maximum target/source length ratio, and its average and standard deviation.}
    \label{tab:pool_stats}
\end{table}

The statistics of each pool for all pairs of languages studied is in \cref{tab:pool_stats}. We observe a similar trend across all language pairs.

\section{Match vs No-match}

\label{app:match-vs-non-match}

\begin{table*}
    \centering
    \footnotesize
    \begin{tabular}{l|cc||cc||cc||cc||cc}
        \multicolumn{11}{c}{En-De} \\\hline
        & \multicolumn{2}{c||}{\texttt{Random}} & \multicolumn{2}{c||}{\texttt{Isometric}} & \multicolumn{2}{c||}{\texttt{Same}} & \multicolumn{2}{c||}{\texttt{Short}} & \multicolumn{2}{c}{\texttt{Tiny}} \\\hline\hline
        Model & No & Yes & No & Yes &  No & Yes & No & Yes & No & Yes \\\hline
        \texttt{gemma2:27b} & 1.100 & 1.097 & 1.099 & 1.094 & 1.098 & 1.097 & \underline{1.087} & \underline{1.011} & \underline{1.066} & \underline{0.955} \\
\texttt{gemma2:9b} & 1.108 & 1.106 & 1.106 & 1.101 & \underline{1.106} & \underline{1.073} & \underline{1.099} & \underline{1.026} & \underline{1.080} & \underline{0.981} \\
\texttt{llama3:70b} & 1.149 & 1.151 & 1.149 & 1.139 & 1.141 & 1.134 & \underline{1.138} & \underline{1.005} & \underline{1.122} & \underline{0.905} \\
\texttt{llama3:8b} & 1.106 & 1.100 & 1.093 & 1.108 & 1.099 & 1.112 & \underline{1.085} & \underline{1.048} & \underline{1.056} & \underline{0.994} \\
\texttt{mistral:7b} & 1.133 & 1.129 & 1.126 & 1.128 & 1.135 & 1.125 & 1.121 & 1.105 & \underline{1.138} & \underline{1.085} \\
\texttt{mixtral:8x7b} & 1.402 & 1.411 & 1.375 & 1.362 & 1.378 & 1.381 & 1.385 & 1.297 & 1.363 & 1.265 \\
\texttt{qwen2:72b} & 1.223 & 1.169 & 1.195 & 1.178 & 1.184 & 1.173 & 1.170 & 1.128 & 1.164 & 1.129 \\
\texttt{qwen2:7b} & 1.132 & 1.160 & 1.144 & 1.125 & 1.129 & 1.129 & 1.128 & 1.135 & 1.117 & 1.095 \\

\multicolumn{11}{c}{} \\
\multicolumn{11}{c}{En-Fr} \\\hline
\texttt{gemma2:27b} & 1.128 & 1.126 & 1.123 & 1.125 & 1.127 & 1.128 & \underline{1.115} & \underline{1.034} & \underline{1.087} & \underline{0.970} \\
\texttt{gemma2:9b} & 1.143 & 1.146 & 1.142 & 1.132 & 1.144 & 1.121 & \underline{1.133} & \underline{1.062} & \underline{1.116} & \underline{1.021} \\
\texttt{llama3:70b} & 1.178 & 1.176 & 1.173 & 1.167 & 1.174 & 1.172 & \underline{1.163} & \underline{1.015} & \underline{1.147} & \underline{0.877} \\
\texttt{llama3:8b} & 1.136 & 1.121 & 1.134 & 1.141 & 1.136 & 1.144 & \underline{1.110} & \underline{1.052} & \underline{1.072} & \underline{0.986} \\
\texttt{mistral:7b} & 1.162 & 1.166 & 1.159 & 1.152 & 1.167 & 1.177 & 1.141 & 1.127 & 1.153 & 1.137 \\
\texttt{mixtral:8x7b} & 1.335 & 1.332 & \underline{1.316} & \underline{1.233} & \underline{1.351} & \underline{1.247} & \underline{1.355} & \underline{1.206} & \underline{1.348} & \underline{1.203} \\
\texttt{qwen2:72b} & 1.178 & 1.180 & 1.198 & 1.174 & 1.184 & 1.171 & \underline{1.215} & \underline{1.139} & \underline{1.172} & \underline{1.118} \\
\texttt{qwen2:7b} & 1.183 & 1.175 & 1.175 & 1.171 & 1.172 & 1.172 & 1.170 & 1.146 & 1.152 & 1.124 \\

\multicolumn{11}{c}{} \\
\multicolumn{11}{c}{En-Es} \\\hline
\texttt{gemma2:27b} & 1.058 & 1.057 & 1.057 & 1.058 & 1.058 & 1.054 & \underline{1.053} & \underline{0.994} & \underline{1.040} & \underline{0.939} \\
\texttt{gemma2:9b} & 1.072 & 1.072 & 1.071 & 1.064 & \underline{1.070} & \underline{1.046} & \underline{1.067} & \underline{1.018} & \underline{1.054} & \underline{0.969} \\
\texttt{llama3:70b} & 1.086 & 1.089 & 1.088 & 1.086 & 1.086 & 1.083 & \underline{1.084} & \underline{0.969} & \underline{1.062} & \underline{0.823} \\
\texttt{llama3:8b} & 1.050 & 1.051 & 1.055 & 1.063 & 1.051 & 1.062 & \underline{1.042} & \underline{1.008} & \underline{0.999} & \underline{0.907} \\
\texttt{mistral:7b} & 1.050 & 1.058 & 1.058 & 1.050 & 1.078 & 1.063 & 1.106 & 1.038 & \underline{1.039} & \underline{1.014} \\
\texttt{mixtral:8x7b} & 1.321 & 1.287 & 1.340 & 1.269 & 1.265 & 1.222 & 1.286 & 1.254 & 1.283 & 1.252 \\
\texttt{qwen2:72b} & 1.116 & 1.113 & 1.108 & 1.110 & 1.113 & 1.117 & 1.114 & 1.091 & 1.106 & 1.075 \\
\texttt{qwen2:7b} & 1.074 & 1.097 & 1.074 & 1.079 & 1.075 & 1.074 & 1.083 & 1.057 & \underline{1.065} & \underline{1.035} \\
    \end{tabular}
    \caption{Average target/source ratios for every pool type when instructions match (`Yes') or do not match (`No') the properties of the samples in 5-shot prompting. Differences for each pool type with $p$-value $< 0.1$ are underlined.}
    \label{tab:difference_match_vs_no_match}
\end{table*}

\begin{figure}[t!]
    \centering
    \includegraphics[width=\linewidth]{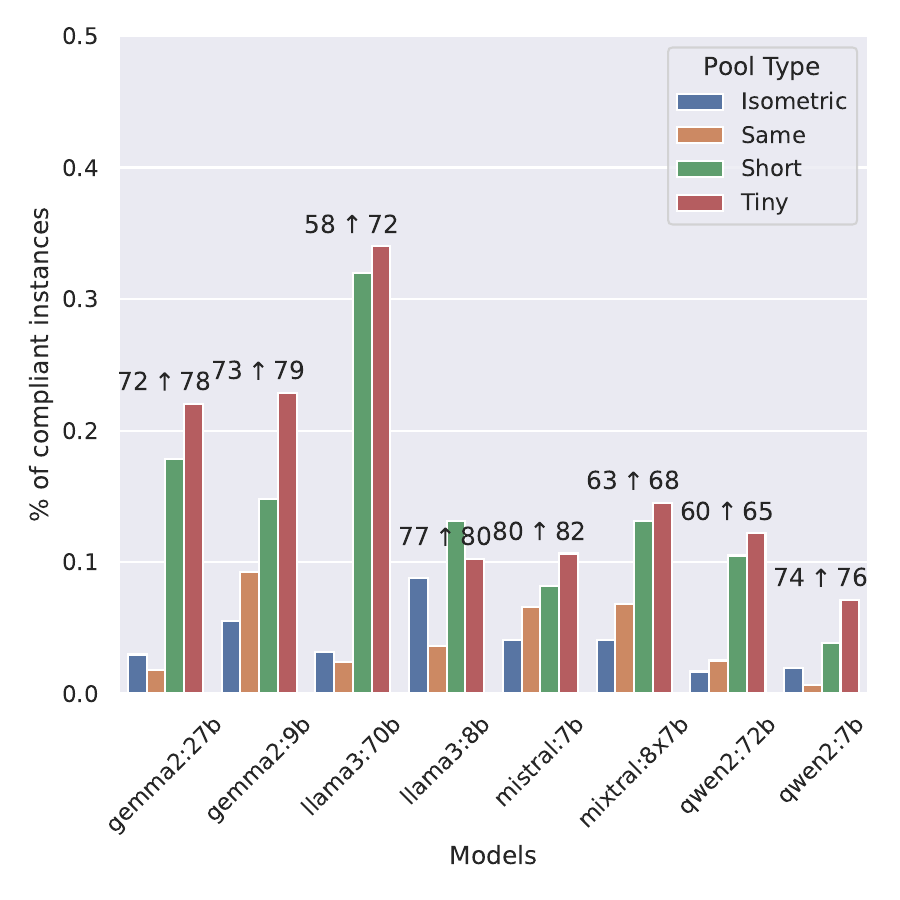}
    \caption{The percentage of input sentences (across all language directions) for which the generated translation meets the isometric condition when the model is prompted to produce isometric, same-length, short, and tiny outputs aligned with respective 5-shot demonstration sets. This evaluation is restricted to input sentences where the particular model did not generate any isometric translation in ten attempts using the uncontrolled prompt.
    }
    \label{fig:comliance_proportion_1_ettempt}
\end{figure}

The comparison of match-vs-non-match for all languages is depicted in \cref{tab:difference_match_vs_no_match}. \cref{fig:comliance_proportion_1_ettempt} shows the proportion of isometric outputs given sentences where each of the models failed to produce isometric translation by default, i.e. under the \emph{Random}-\texttt{Random} setting, even across 10 default runs. The translations are taken after only one attempt, which is in contrast to \cref{fig:comliance_proportion} where the outputs are selected from 10 attempts.


\section{Few-shot Prompting}
\label{app:few-shot}

\begin{figure*}[t]
    \centering
    \includegraphics[width=\linewidth]{figures/ratio.pool_types.en-de.common-range.pdf}
    \includegraphics[width=\linewidth]{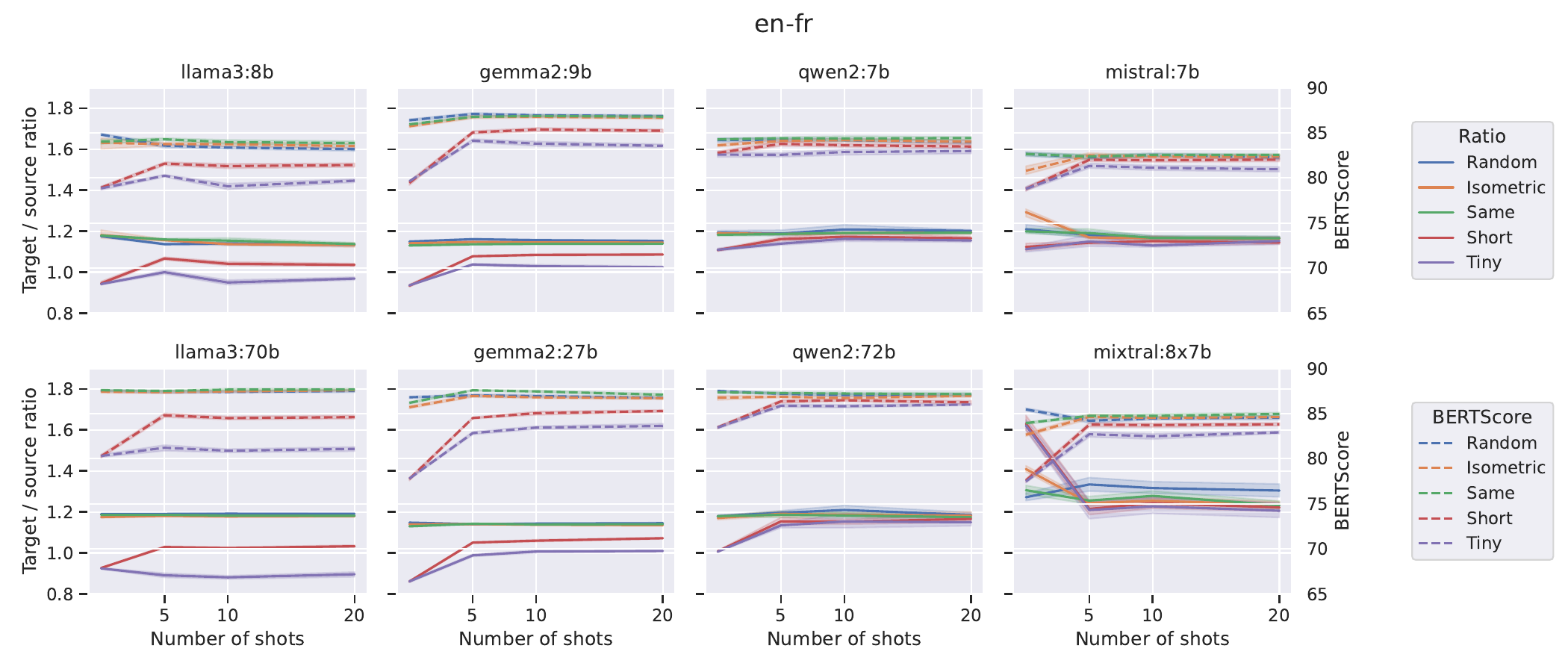}
    \includegraphics[width=\linewidth]{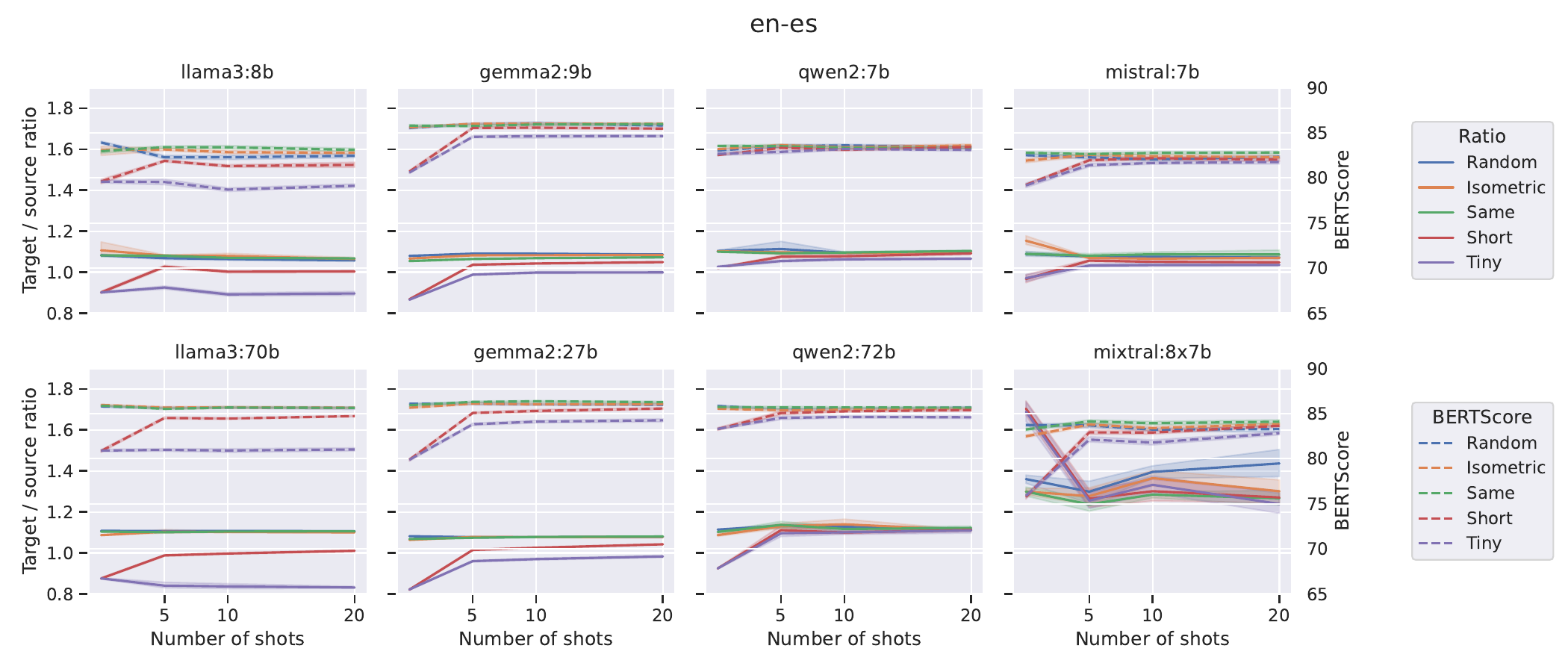}
    \caption{The translation quality (BERTScore, dashed lines and the right hand y-axes) and length ratio (solid lines and left-hand y-axes) for all few-shot settings, models and language pairs.}
    \label{fig:bs_ratio_pools}
\end{figure*}

The comparison between all few shot settings for all languages is displayed in \cref{fig:bs_ratio_pools}. Additionally, we provide a more detailed view of the results of all few-shot settings, which is presented in \cref{tab:pool_type_evaluation_0} (zero-shot), \cref{tab:pool_type_evaluation_5} (5-shot), \cref{tab:pool_type_evaluation_10} (10-shot) and \cref{tab:pool_type_evaluation_20} (20-shot). We also compare these results to an oracle setup, in which the best translation is selected based on the sentence BLEU score across all configurations, after filtering out translations that do not fall within ±10\% of the source character count.

\section{Translation Quality and Length Tradeoff}
\label{app:tradeoff}

\begin{figure*}
    \centering
    \includegraphics[width=\linewidth]{figures/tradeoff.models.en-de.pdf}
    \includegraphics[width=\linewidth]{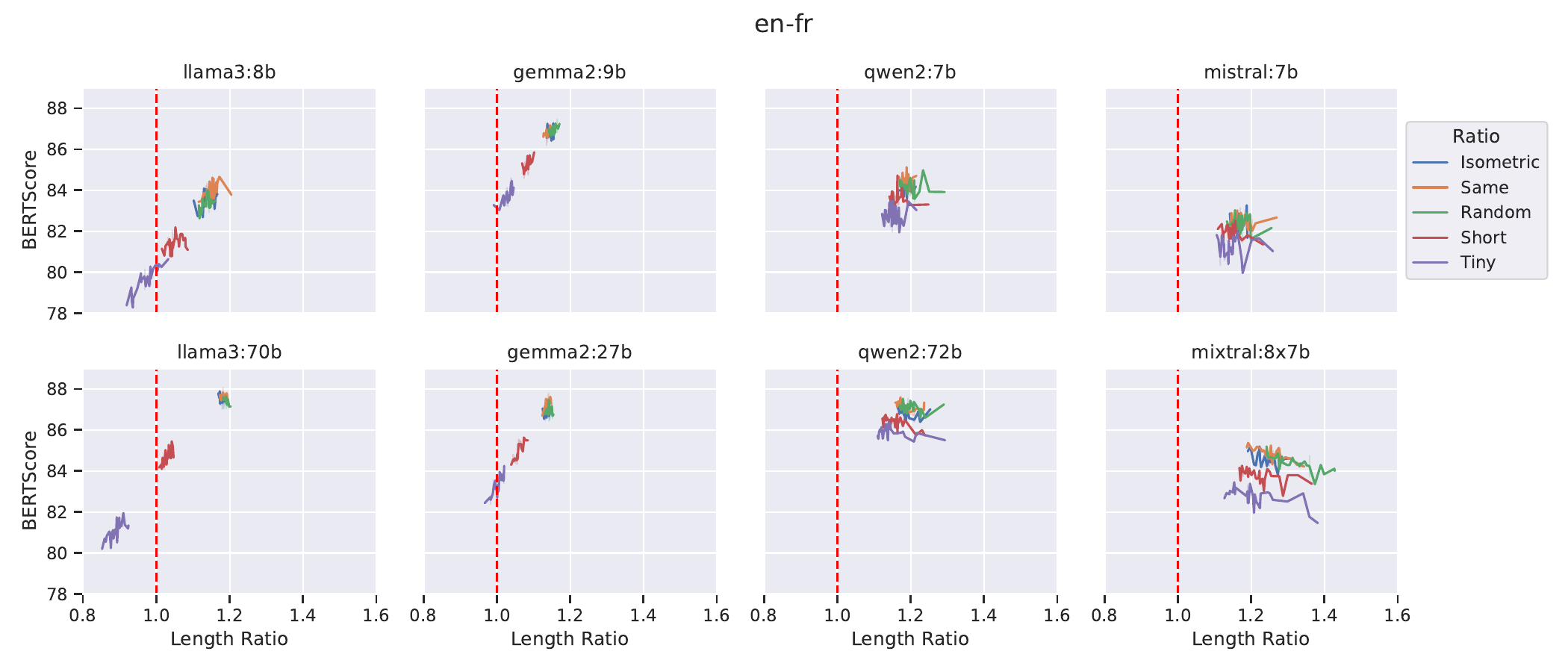}
    \includegraphics[width=\linewidth]{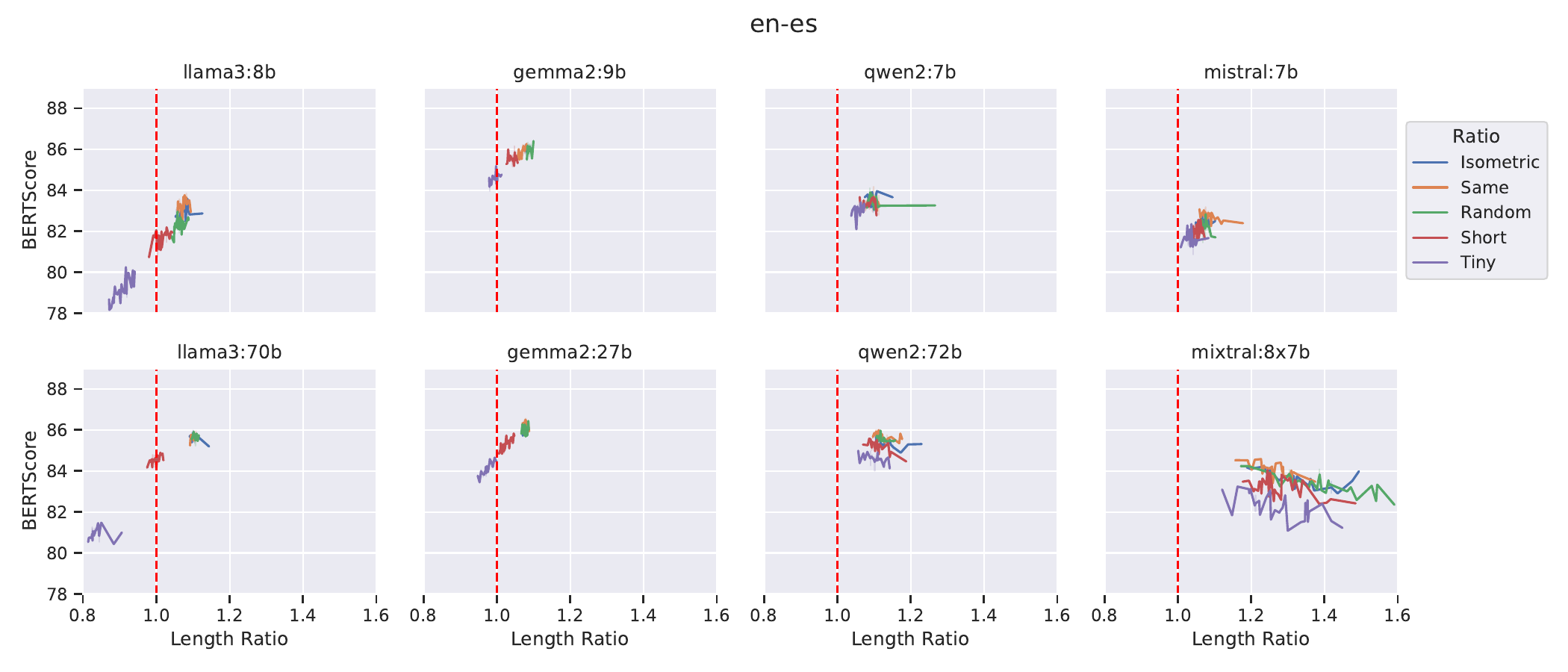}
    \caption{Trade-off between the length ratio (x-axis) and translation quality (y-axis) for 5, 10, 20-shot settings and all models and language pairs.}
    \label{fig:tradeoff}
\end{figure*}

The length ratio and translation quality tradeoff for all languages is presented in \cref{fig:tradeoff}. We observe that models generally produce isometric translation when \emph{Tiny} setting is used. The exception is Spanish, where the average 1.0 length ratio can be obtained by \emph{Short} setting. This is in line with our intuition since Spanish exhibits a smaller length ratio of 1.04 for the training data from MuST-C, compared to length ratios of 1.12 and 1.11 for German and French, respectively.\footnote{These values were calculated by the organizers of the isometric shared task and are mentioned on the official website \url{https://iwslt.org/2022/isometric}.}

\begin{table*}[t]
    \centering
    \footnotesize
    \setlength{\tabcolsep}{4pt}
\begin{tabular}{ll|cccc|cccc|cccc}
&& \multicolumn{4}{c|}{En-De} & \multicolumn{4}{c|}{En-Fr} & \multicolumn{4}{c}{En-Es} \\\hline
Model & Prompt Type & LR & LC & BS & BLEU & LR & LC & BS & BLEU & LR & LC & BS & BLEU \\
\hline\hline
 & \texttt{Random} & 1.13 & 37.39 & 83.57 & 31.82 & 1.15 & 37.78 & 86.80 & 43.29 & 1.08 & 48.20 & 86.08 & 39.24 \\
\multirow[c]{2}{*}{\texttt{gemma2:27b}} & \texttt{Isometric} & 1.12 & 39.10 & 83.28 & 29.94 & 1.14 & 39.10 & 85.70 & 38.25 & 1.06 & 51.65 & 85.65 & 37.54 \\
 & \texttt{Same} & 1.12 & 41.55 & 83.52 & 31.02 & 1.13 & 44.45 & 86.18 & 40.07 & 1.07 & 51.95 & 85.92 & 37.75 \\
 & \texttt{Short} & 0.83 & 27.05 & 76.10 & 12.16 & 0.86 & 33.85 & 77.74 & 15.91 & 0.82 & 30.25 & 79.88 & 19.15 \\
\cline{1-14}
 & \texttt{Random} & 1.13 & 34.85 & 83.43 & 31.47 & 1.15 & 36.75 & 86.40 & 41.81 & 1.08 & 48.85 & 85.54 & 37.87 \\
\multirow[c]{2}{*}{\texttt{gemma2:9b}} & \texttt{Isometric} & 1.12 & 38.75 & 82.93 & 29.37 & 1.14 & 38.70 & 85.73 & 38.18 & 1.07 & 54.40 & 85.60 & 36.59 \\
 & \texttt{Same} & 1.11 & 41.45 & 82.89 & 29.63 & 1.13 & 40.75 & 85.94 & 39.55 & 1.05 & 55.95 & 85.80 & 36.91 \\
 & \texttt{Short} & 0.91 & 35.65 & 78.15 & 16.12 & 0.93 & 38.65 & 79.45 & 20.16 & 0.87 & 36.45 & 80.70 & 21.66 \\
\cline{1-14}
 & \texttt{Random} & 1.18 & 26.94 & 84.58 & 34.31 & 1.19 & 28.06 & 87.52 & 44.24 & 1.11 & 45.83 & 85.78 & 37.08 \\
\multirow[c]{2}{*}{\texttt{llama3:70b}} & \texttt{Isometric} & 1.16 & 31.80 & 84.53 & 33.79 & 1.17 & 31.40 & 87.42 & 43.65 & 1.09 & 49.65 & 85.95 & 37.36 \\
 & \texttt{Same} & 1.17 & 26.70 & 84.75 & 34.87 & 1.19 & 28.35 & 87.58 & 44.24 & 1.10 & 46.95 & 85.87 & 36.72 \\
 & \texttt{Short} & 0.92 & 38.75 & 79.04 & 17.20 & 0.93 & 38.40 & 80.26 & 21.27 & 0.88 & 37.05 & 80.85 & 23.17 \\
\cline{1-14}
 & \texttt{Random} & 1.16 & 31.60 & 82.70 & 28.96 & 1.18 & 31.39 & 84.81 & 36.68 & 1.08 & 50.10 & 83.93 & 32.78 \\
\multirow[c]{2}{*}{\texttt{llama3:8b}} & \texttt{Isometric} & 1.18 & 29.33 & 82.23 & 27.82 & 1.18 & 29.67 & 83.90 & 34.23 & 1.11 & 47.00 & 83.10 & 31.00 \\
 & \texttt{Same} & 1.17 & 28.65 & 82.55 & 28.36 & 1.18 & 27.50 & 84.01 & 34.70 & 1.08 & 47.33 & 82.96 & 30.99 \\
 & \texttt{Short} & 0.96 & 33.85 & 77.88 & 15.83 & 0.95 & 37.10 & 78.92 & 18.29 & 0.90 & 39.70 & 79.61 & 19.69 \\
\cline{1-14}
 & \texttt{Random} & 1.20 & 30.40 & 77.28 & 23.33 & 1.21 & 32.40 & 82.65 & 30.19 & 1.09 & 48.70 & 82.52 & 29.49 \\
\multirow[c]{2}{*}{\texttt{mistral:7b}} & \texttt{Isometric} & 1.29 & 24.28 & 72.26 & 19.88 & 1.29 & 26.67 & 80.79 & 28.20 & 1.15 & 43.89 & 81.92 & 28.57 \\
 & \texttt{Same} & 1.18 & 28.40 & 80.55 & 23.23 & 1.20 & 32.90 & 82.66 & 29.99 & 1.09 & 47.40 & 82.79 & 29.30 \\
 & \texttt{Short} & 1.13 & 39.40 & 73.62 & 16.99 & 1.12 & 41.30 & 78.78 & 23.39 & 0.97 & 45.55 & 79.23 & 23.13 \\
\cline{1-14}
 & \texttt{Random} & 1.42 & 23.75 & 81.50 & 29.00 & 1.27 & 26.00 & 85.45 & 38.28 & 1.36 & 38.83 & 83.72 & 31.61 \\
\multirow[c]{2}{*}{\texttt{mixtral:8x7b}} & \texttt{Isometric} & 1.49 & 15.89 & 79.81 & 26.21 & 1.41 & 18.85 & 82.65 & 34.47 & 1.30 & 29.50 & 82.48 & 30.23 \\
 & \texttt{Same} & 1.53 & 21.50 & 79.77 & 26.86 & 1.31 & 28.78 & 83.94 & 36.18 & 1.30 & 39.45 & 83.24 & 30.73 \\
 & \texttt{Short} & 1.81 & 18.50 & 73.92 & 18.32 & 1.63 & 28.30 & 77.58 & 26.18 & 1.70 & 26.30 & 75.75 & 20.91 \\
\cline{1-14}
 & \texttt{Random} & 1.17 & 30.28 & 84.02 & 33.07 & 1.18 & 33.61 & 87.50 & 44.30 & 1.11 & 45.44 & 85.83 & 37.82 \\
\multirow[c]{2}{*}{\texttt{qwen2:72b}} & \texttt{Isometric} & 1.15 & 35.65 & 83.69 & 31.80 & 1.17 & 38.15 & 86.76 & 41.92 & 1.09 & 50.60 & 85.53 & 36.41 \\
 & \texttt{Same} & 1.16 & 32.50 & 83.62 & 32.22 & 1.18 & 38.45 & 87.35 & 44.87 & 1.10 & 50.70 & 85.73 & 36.86 \\
 & \texttt{Short} & 0.98 & 40.10 & 80.48 & 20.71 & 1.01 & 43.95 & 83.46 & 27.91 & 0.92 & 37.60 & 83.27 & 25.77 \\
\cline{1-14}
 & \texttt{Random} & 1.21 & 28.55 & 80.83 & 24.32 & 1.19 & 28.81 & 84.20 & 36.04 & 1.10 & 41.45 & 83.04 & 30.98 \\
\multirow[c]{2}{*}{\texttt{qwen2:7b}} & \texttt{Isometric} & 1.20 & 27.50 & 80.51 & 23.05 & 1.19 & 30.75 & 83.61 & 34.00 & 1.10 & 45.72 & 83.19 & 30.33 \\
 & \texttt{Same} & 1.19 & 28.45 & 81.04 & 23.67 & 1.18 & 29.80 & 84.27 & 35.42 & 1.10 & 43.90 & 83.55 & 30.91 \\
 & \texttt{Short} & 1.09 & 38.00 & 79.34 & 18.89 & 1.11 & 41.55 & 82.78 & 30.33 & 1.02 & 49.65 & 82.54 & 27.35 \\\hline
 Oracle &  & 1.07 & 65.00 & 87.74 & 49.60 & 1.05 & 76.50 & 87.99 & 55.60 & 1.03 & 83.00 & 88.47 & 53.50 \\\hline
\end{tabular}
    \caption{0-shot prompting for all language pairs. Columns denote length ratio (LR), length compliance (LC), BERTScore (BS) and BLEU.}
    \label{tab:pool_type_evaluation_0}
\end{table*}

\begin{table*}[t]
    \centering
    \footnotesize
    \setlength{\tabcolsep}{4pt}
\begin{tabular}{ll|cccc|cccc|cccc}
&& \multicolumn{4}{c|}{En-De} & \multicolumn{4}{c|}{En-Fr} & \multicolumn{4}{c}{En-Es} \\\hline
Model & Pool Type & LR & LC & BS & BLEU & LR & LC & BS & BLEU & LR & LC & BS & BLEU \\
\hline\hline
 & \texttt{Random} & 1.13 & 36.75 & 83.97 & 32.81 & 1.14 & 39.70 & 87.01 & 42.86 & 1.08 & 50.60 & 86.11 & 39.02 \\
\multirow[c]{3}{*}{\texttt{gemma2:27b}} & \texttt{Isometric} & 1.13 & 39.15 & 84.04 & 33.07 & 1.14 & 40.35 & 86.95 & 42.29 & 1.08 & 51.55 & 86.10 & 38.61 \\
 & \texttt{Same} & 1.13 & 38.65 & 84.18 & 33.05 & 1.14 & 40.35 & 87.58 & 43.81 & 1.07 & 52.50 & 86.29 & 38.71 \\
 & \texttt{Short} & 1.04 & 44.05 & 82.41 & 28.01 & 1.05 & 42.65 & 84.50 & 34.61 & 1.01 & 53.15 & 85.06 & 35.54 \\
 & \texttt{Tiny} & 0.99 & 41.00 & 81.13 & 24.46 & 0.99 & 42.55 & 82.83 & 30.89 & 0.96 & 48.05 & 83.81 & 32.21 \\
\cline{1-14}
 & \texttt{Random} & 1.14 & 35.35 & 84.07 & 32.37 & 1.16 & 37.85 & 87.09 & 43.13 & 1.09 & 50.90 & 85.99 & 37.59 \\
\multirow[c]{3}{*}{\texttt{gemma2:9b}} & \texttt{Isometric} & 1.14 & 37.25 & 83.93 & 31.80 & 1.15 & 40.30 & 86.76 & 42.09 & 1.08 & 51.85 & 86.00 & 37.59 \\
 & \texttt{Same} & 1.11 & 41.35 & 83.72 & 30.57 & 1.14 & 43.00 & 86.77 & 42.05 & 1.06 & 56.50 & 85.75 & 37.09 \\
 & \texttt{Short} & 1.06 & 44.25 & 82.71 & 28.24 & 1.08 & 46.50 & 85.05 & 36.13 & 1.04 & 56.60 & 85.54 & 36.71 \\
 & \texttt{Tiny} & 1.01 & 44.50 & 81.55 & 25.40 & 1.04 & 44.10 & 84.13 & 34.59 & 0.99 & 53.80 & 84.55 & 33.53 \\
\cline{1-14}
 & \texttt{Random} & 1.19 & 26.45 & 84.85 & 34.86 & 1.19 & 29.15 & 87.40 & 44.08 & 1.11 & 48.45 & 85.62 & 36.89 \\
\multirow[c]{3}{*}{\texttt{llama3:70b}} & \texttt{Isometric} & 1.17 & 27.90 & 84.65 & 34.25 & 1.18 & 30.85 & 87.37 & 43.63 & 1.10 & 47.90 & 85.66 & 37.07 \\
 & \texttt{Same} & 1.17 & 28.50 & 84.83 & 34.84 & 1.19 & 31.65 & 87.50 & 43.96 & 1.10 & 49.60 & 85.52 & 36.43 \\
 & \texttt{Short} & 1.04 & 40.90 & 82.76 & 28.12 & 1.03 & 44.00 & 84.81 & 35.36 & 0.99 & 50.15 & 84.51 & 33.68 \\
 & \texttt{Tiny} & 0.94 & 35.80 & 80.61 & 22.66 & 0.89 & 37.05 & 81.20 & 25.79 & 0.84 & 33.75 & 80.97 & 24.73 \\
\cline{1-14}
 & \texttt{Random} & 1.14 & 32.00 & 81.55 & 26.44 & 1.14 & 34.22 & 83.57 & 34.32 & 1.07 & 48.70 & 82.29 & 30.53 \\
\multirow[c]{3}{*}{\texttt{llama3:8b}} & \texttt{Isometric} & 1.15 & 32.00 & 81.87 & 26.57 & 1.16 & 36.30 & 83.74 & 34.08 & 1.08 & 49.40 & 83.16 & 31.45 \\
 & \texttt{Same} & 1.15 & 33.40 & 82.06 & 27.04 & 1.16 & 34.50 & 84.28 & 34.80 & 1.08 & 51.45 & 83.41 & 31.05 \\
 & \texttt{Short} & 1.08 & 39.05 & 80.69 & 23.71 & 1.07 & 38.55 & 81.59 & 29.35 & 1.03 & 51.40 & 81.88 & 28.56 \\
 & \texttt{Tiny} & 1.03 & 37.55 & 79.41 & 21.21 & 1.00 & 38.10 & 80.24 & 26.74 & 0.93 & 43.15 & 79.55 & 23.71 \\
\cline{1-14}
 & \texttt{Random} & 1.17 & 32.00 & 80.66 & 23.66 & 1.18 & 35.20 & 82.29 & 29.70 & 1.08 & 48.78 & 82.28 & 28.52 \\
\multirow[c]{3}{*}{\texttt{mistral:7b}} & \texttt{Isometric} & 1.17 & 32.55 & 80.76 & 23.67 & 1.17 & 37.40 & 82.50 & 30.11 & 1.07 & 50.80 & 82.57 & 28.79 \\
 & \texttt{Same} & 1.16 & 33.50 & 80.93 & 23.86 & 1.19 & 34.90 & 82.37 & 29.91 & 1.08 & 52.60 & 82.64 & 28.99 \\
 & \texttt{Short} & 1.14 & 38.15 & 80.30 & 22.68 & 1.14 & 39.10 & 82.01 & 29.06 & 1.06 & 51.85 & 81.96 & 28.64 \\
 & \texttt{Tiny} & 1.12 & 38.65 & 79.75 & 21.46 & 1.15 & 40.00 & 81.34 & 28.10 & 1.03 & 52.25 & 81.42 & 27.36 \\
\cline{1-14}
 & \texttt{Random} & 1.43 & 26.85 & 81.99 & 30.58 & 1.33 & 30.00 & 84.20 & 38.51 & 1.30 & 40.45 & 83.68 & 33.44 \\
\multirow[c]{3}{*}{\texttt{mixtral:8x7b}} & \texttt{Isometric} & 1.38 & 29.20 & 82.02 & 30.71 & 1.24 & 32.10 & 84.64 & 38.40 & 1.28 & 42.80 & 83.78 & 33.85 \\
 & \texttt{Same} & 1.40 & 32.40 & 82.38 & 31.10 & 1.26 & 32.35 & 84.74 & 39.20 & 1.24 & 44.45 & 84.12 & 34.18 \\
 & \texttt{Short} & 1.32 & 33.65 & 81.21 & 28.83 & 1.21 & 37.45 & 83.77 & 35.39 & 1.26 & 45.75 & 82.91 & 32.28 \\
 & \texttt{Tiny} & 1.29 & 36.80 & 79.98 & 26.16 & 1.21 & 38.56 & 82.70 & 33.03 & 1.25 & 46.80 & 82.11 & 31.02 \\
\cline{1-14}
 & \texttt{Random} & 1.20 & 29.25 & 84.08 & 33.25 & 1.20 & 33.35 & 87.18 & 43.74 & 1.13 & 45.65 & 85.48 & 37.00 \\
\multirow[c]{3}{*}{\texttt{qwen2:72b}} & \texttt{Isometric} & 1.21 & 29.95 & 83.98 & 33.02 & 1.19 & 34.60 & 86.85 & 42.69 & 1.13 & 46.10 & 85.37 & 36.77 \\
 & \texttt{Same} & 1.21 & 31.20 & 83.98 & 32.91 & 1.19 & 35.80 & 87.26 & 43.57 & 1.14 & 48.60 & 85.66 & 37.52 \\
 & \texttt{Short} & 1.16 & 34.40 & 83.63 & 31.59 & 1.15 & 41.60 & 86.36 & 40.17 & 1.11 & 49.65 & 85.05 & 36.10 \\
 & \texttt{Tiny} & 1.16 & 35.75 & 83.03 & 30.09 & 1.13 & 42.80 & 85.86 & 38.80 & 1.10 & 49.40 & 84.52 & 35.20 \\
\cline{1-14}
 & \texttt{Random} & 1.20 & 32.10 & 81.04 & 23.99 & 1.19 & 31.75 & 84.24 & 35.37 & 1.11 & 47.40 & 83.53 & 30.89 \\
\multirow[c]{3}{*}{\texttt{qwen2:7b}} & \texttt{Isometric} & 1.16 & 30.30 & 81.24 & 23.78 & 1.19 & 32.30 & 84.11 & 34.86 & 1.10 & 48.70 & 83.63 & 31.13 \\
 & \texttt{Same} & 1.17 & 31.55 & 81.28 & 24.07 & 1.19 & 32.80 & 84.41 & 35.16 & 1.09 & 47.75 & 83.52 & 30.98 \\
 & \texttt{Short} & 1.14 & 35.56 & 80.97 & 22.99 & 1.16 & 35.25 & 83.77 & 34.09 & 1.08 & 48.95 & 83.33 & 30.75 \\
 & \texttt{Tiny} & 1.13 & 34.89 & 80.63 & 22.43 & 1.14 & 36.60 & 82.56 & 32.26 & 1.05 & 50.00 & 82.89 & 29.95 \\\hline
 Oracle &  & 1.07 & 65.00 & 87.74 & 49.60 & 1.05 & 76.50 & 87.99 & 55.60 & 1.03 & 83.00 & 88.47 & 53.50 \\\hline
\end{tabular}
    \caption{5-shot prompting for all language pairs when sampling examples from different pools. Columns denote length ratio (LR), length compliance (LC), BERTScore (BS) and BLEU. All numbers are averaged across 10 instances. The prompt text \emph{matches} the pool type. }
    \label{tab:pool_type_evaluation_5}
\end{table*}

\begin{table*}[t]
    \centering
    \footnotesize
    \setlength{\tabcolsep}{4pt}
\begin{tabular}{ll|cccc|cccc|cccc}
&& \multicolumn{4}{c|}{En-De} & \multicolumn{4}{c|}{En-Fr} & \multicolumn{4}{c}{En-Es} \\\hline
Model & Pool Type & LR & LC & BS & BLEU & LR & LC & BS & BLEU & LR & LC & BS & BLEU \\
\hline\hline
 & \texttt{Random} & 1.13 & 37.20 & 84.14 & 33.34 & 1.14 & 41.50 & 86.93 & 42.54 & 1.08 & 50.65 & 86.01 & 38.58 \\
\multirow[c]{3}{*}{\texttt{gemma2:27b}} & \texttt{Isometric} & 1.13 & 39.10 & 83.98 & 32.89 & 1.14 & 40.35 & 86.79 & 41.72 & 1.08 & 50.55 & 86.03 & 38.50 \\
 & \texttt{Same} & 1.13 & 38.80 & 84.19 & 33.31 & 1.14 & 41.70 & 87.45 & 43.59 & 1.08 & 50.90 & 86.36 & 39.01 \\
 & \texttt{Short} & 1.05 & 43.95 & 82.76 & 28.38 & 1.06 & 45.40 & 85.04 & 36.26 & 1.02 & 52.15 & 85.29 & 36.17 \\
 & \texttt{Tiny} & 1.00 & 41.05 & 81.25 & 24.84 & 1.01 & 41.55 & 83.44 & 33.10 & 0.97 & 47.70 & 84.10 & 33.15 \\
\cline{1-14}
 & \texttt{Random} & 1.14 & 36.75 & 83.91 & 32.23 & 1.16 & 39.85 & 86.95 & 43.22 & 1.09 & 50.35 & 86.05 & 37.93 \\
\multirow[c]{3}{*}{\texttt{gemma2:9b}} & \texttt{Isometric} & 1.13 & 36.35 & 83.89 & 31.57 & 1.15 & 41.40 & 86.76 & 41.89 & 1.08 & 53.05 & 86.00 & 37.68 \\
 & \texttt{Same} & 1.12 & 40.15 & 83.80 & 30.88 & 1.14 & 42.45 & 86.86 & 42.48 & 1.07 & 55.40 & 85.93 & 37.53 \\
 & \texttt{Short} & 1.07 & 43.55 & 82.81 & 28.84 & 1.08 & 47.35 & 85.36 & 37.34 & 1.04 & 58.45 & 85.57 & 36.41 \\
 & \texttt{Tiny} & 1.03 & 43.85 & 81.92 & 26.68 & 1.03 & 43.95 & 83.80 & 34.25 & 1.00 & 55.35 & 84.63 & 33.89 \\
\cline{1-14}
 & \texttt{Random} & 1.19 & 26.30 & 84.87 & 35.16 & 1.19 & 30.10 & 87.39 & 44.23 & 1.11 & 47.15 & 85.67 & 36.88 \\
\multirow[c]{3}{*}{\texttt{llama3:70b}} & \texttt{Isometric} & 1.18 & 27.70 & 84.79 & 34.74 & 1.18 & 30.75 & 87.45 & 43.84 & 1.10 & 46.75 & 85.68 & 36.90 \\
 & \texttt{Same} & 1.17 & 28.90 & 85.00 & 35.04 & 1.18 & 31.45 & 87.65 & 44.37 & 1.10 & 48.95 & 85.67 & 36.73 \\
 & \texttt{Short} & 1.05 & 40.00 & 82.94 & 28.86 & 1.02 & 42.45 & 84.49 & 34.89 & 1.00 & 50.05 & 84.44 & 34.19 \\
 & \texttt{Tiny} & 0.95 & 36.10 & 81.01 & 23.37 & 0.88 & 36.50 & 80.87 & 25.85 & 0.84 & 34.95 & 80.90 & 25.09 \\
\cline{1-14}
 & \texttt{Random} & 1.14 & 33.30 & 81.07 & 25.66 & 1.14 & 34.22 & 83.37 & 33.56 & 1.06 & 49.20 & 82.29 & 30.24 \\
\multirow[c]{3}{*}{\texttt{llama3:8b}} & \texttt{Isometric} & 1.14 & 33.20 & 81.77 & 26.21 & 1.14 & 36.06 & 83.75 & 34.38 & 1.08 & 49.55 & 82.85 & 31.05 \\
 & \texttt{Same} & 1.15 & 33.80 & 81.88 & 26.53 & 1.15 & 34.05 & 83.97 & 34.57 & 1.07 & 52.00 & 83.40 & 31.27 \\
 & \texttt{Short} & 1.06 & 36.35 & 80.16 & 22.92 & 1.04 & 39.60 & 81.31 & 29.40 & 1.00 & 48.00 & 81.31 & 27.55 \\
 & \texttt{Tiny} & 1.00 & 35.75 & 78.99 & 20.19 & 0.95 & 35.90 & 79.07 & 24.33 & 0.89 & 42.15 & 78.70 & 22.47 \\
\cline{1-14}
 & \texttt{Random} & 1.16 & 32.40 & 80.52 & 23.18 & 1.17 & 34.33 & 82.56 & 30.17 & 1.08 & 47.65 & 82.06 & 28.27 \\
\multirow[c]{3}{*}{\texttt{mistral:7b}} & \texttt{Isometric} & 1.15 & 32.95 & 80.79 & 23.87 & 1.17 & 36.83 & 82.36 & 29.99 & 1.07 & 49.60 & 82.39 & 28.73 \\
 & \texttt{Same} & 1.16 & 33.45 & 80.85 & 23.59 & 1.17 & 36.40 & 82.54 & 30.25 & 1.09 & 50.20 & 82.77 & 29.37 \\
 & \texttt{Short} & 1.13 & 37.10 & 80.45 & 22.91 & 1.15 & 38.35 & 81.97 & 29.10 & 1.05 & 52.05 & 82.19 & 28.43 \\
 & \texttt{Tiny} & 1.13 & 37.85 & 79.66 & 21.53 & 1.13 & 41.25 & 81.13 & 28.11 & 1.04 & 51.40 & 81.65 & 27.74 \\
\cline{1-14}
 & \texttt{Random} & 1.46 & 28.15 & 82.02 & 30.81 & 1.32 & 31.60 & 84.45 & 38.81 & 1.40 & 40.35 & 83.22 & 33.38 \\
\multirow[c]{3}{*}{\texttt{mixtral:8x7b}} & \texttt{Isometric} & 1.35 & 31.95 & 82.21 & 31.09 & 1.25 & 31.75 & 84.52 & 38.81 & 1.36 & 42.15 & 83.37 & 33.56 \\
 & \texttt{Same} & 1.42 & 30.95 & 81.95 & 30.69 & 1.28 & 33.75 & 84.73 & 38.92 & 1.28 & 43.20 & 83.92 & 34.50 \\
 & \texttt{Short} & 1.36 & 35.75 & 81.16 & 28.92 & 1.24 & 37.15 & 83.72 & 35.97 & 1.30 & 46.45 & 82.89 & 32.66 \\
 & \texttt{Tiny} & 1.27 & 39.20 & 80.21 & 26.27 & 1.23 & 37.80 & 82.48 & 32.84 & 1.33 & 45.00 & 81.78 & 31.69 \\
\cline{1-14}
 & \texttt{Random} & 1.22 & 29.50 & 84.12 & 33.35 & 1.21 & 34.40 & 87.05 & 43.42 & 1.13 & 46.15 & 85.52 & 37.22 \\
\multirow[c]{3}{*}{\texttt{qwen2:72b}} & \texttt{Isometric} & 1.20 & 28.70 & 84.11 & 33.29 & 1.19 & 36.00 & 86.73 & 42.56 & 1.14 & 45.60 & 85.42 & 36.82 \\
 & \texttt{Same} & 1.22 & 31.35 & 83.97 & 32.99 & 1.18 & 37.05 & 87.22 & 43.57 & 1.12 & 48.10 & 85.68 & 37.43 \\
 & \texttt{Short} & 1.14 & 35.30 & 83.65 & 31.82 & 1.15 & 42.45 & 86.47 & 40.71 & 1.10 & 49.10 & 85.24 & 36.88 \\
 & \texttt{Tiny} & 1.14 & 35.55 & 83.20 & 30.38 & 1.15 & 43.55 & 85.81 & 39.52 & 1.10 & 50.15 & 84.62 & 35.44 \\
\cline{1-14}
 & \texttt{Random} & 1.18 & 29.10 & 81.17 & 23.82 & 1.21 & 31.65 & 84.19 & 35.02 & 1.09 & 46.06 & 83.61 & 30.71 \\
\multirow[c]{3}{*}{\texttt{qwen2:7b}} & \texttt{Isometric} & 1.17 & 32.05 & 81.31 & 23.71 & 1.19 & 31.20 & 84.19 & 35.11 & 1.09 & 45.35 & 83.40 & 30.67 \\
 & \texttt{Same} & 1.17 & 31.20 & 81.37 & 23.99 & 1.19 & 32.00 & 84.36 & 35.03 & 1.10 & 45.95 & 83.33 & 30.93 \\
 & \texttt{Short} & 1.15 & 33.55 & 80.81 & 23.13 & 1.17 & 35.00 & 83.63 & 34.17 & 1.08 & 47.00 & 83.12 & 30.45 \\
 & \texttt{Tiny} & 1.13 & 37.00 & 80.55 & 22.31 & 1.16 & 34.30 & 82.87 & 33.13 & 1.06 & 49.00 & 83.20 & 30.41 \\\hline
 Oracle &  & 1.07 & 65.00 & 87.74 & 49.60 & 1.05 & 76.50 & 87.99 & 55.60 & 1.03 & 83.00 & 88.47 & 53.50 \\\hline
\end{tabular}
    \caption{10-shot prompting for all language pairs when sampling examples from different pools. Columns denote length ratio (LR), length compliance (LC), BERTScore (BS) and BLEU. All numbers are averaged across 10 instances. The prompt text \emph{matches} the pool type. }
    \label{tab:pool_type_evaluation_10}
\end{table*}

\begin{table*}[t]
    \centering
    \footnotesize
    \setlength{\tabcolsep}{4pt}
\begin{tabular}{ll|cccc|cccc|cccc}
&& \multicolumn{4}{c|}{En-De} & \multicolumn{4}{c|}{En-Fr} & \multicolumn{4}{c}{En-Es} \\\hline
Model & Pool Type & LR & LC & BS & BLEU & LR & LC & BS & BLEU & LR & LC & BS & BLEU \\
\hline\hline
 & \texttt{Random} & 1.13 & 38.05 & 84.15 & 33.75 & 1.14 & 39.40 & 86.74 & 42.23 & 1.08 & 49.35 & 85.98 & 38.82 \\
\multirow[c]{3}{*}{\texttt{gemma2:27b}} & \texttt{Isometric} & 1.13 & 39.70 & 84.08 & 33.24 & 1.14 & 39.60 & 86.71 & 42.21 & 1.08 & 48.90 & 86.05 & 38.70 \\
 & \texttt{Same} & 1.13 & 39.15 & 84.29 & 33.68 & 1.14 & 42.50 & 87.08 & 42.86 & 1.08 & 49.85 & 86.26 & 39.03 \\
 & \texttt{Short} & 1.07 & 43.45 & 83.07 & 30.32 & 1.07 & 45.05 & 85.28 & 37.52 & 1.04 & 52.20 & 85.55 & 37.43 \\
 & \texttt{Tiny} & 1.01 & 42.00 & 81.88 & 27.16 & 1.01 & 40.90 & 83.62 & 33.65 & 0.98 & 48.55 & 84.25 & 34.32 \\
\cline{1-14}
 & \texttt{Random} & 1.14 & 37.30 & 83.81 & 32.08 & 1.15 & 39.60 & 86.86 & 43.01 & 1.09 & 50.72 & 85.82 & 37.94 \\
\multirow[c]{3}{*}{\texttt{gemma2:9b}} & \texttt{Isometric} & 1.13 & 37.75 & 83.73 & 31.51 & 1.14 & 41.10 & 86.68 & 42.17 & 1.08 & 51.40 & 86.01 & 37.93 \\
 & \texttt{Same} & 1.12 & 39.80 & 83.72 & 30.97 & 1.14 & 42.40 & 86.79 & 42.11 & 1.07 & 54.05 & 85.97 & 38.00 \\
 & \texttt{Short} & 1.08 & 41.20 & 82.83 & 28.93 & 1.09 & 46.70 & 85.23 & 37.48 & 1.05 & 54.85 & 85.46 & 36.86 \\
 & \texttt{Tiny} & 1.04 & 46.70 & 81.94 & 26.43 & 1.02 & 43.85 & 83.56 & 33.87 & 1.00 & 54.35 & 84.64 & 33.62 \\
\cline{1-14}
 & \texttt{Random} & 1.19 & 27.50 & 85.01 & 35.32 & 1.19 & 30.95 & 87.50 & 44.27 & 1.11 & 47.80 & 85.61 & 36.93 \\
\multirow[c]{3}{*}{\texttt{llama3:70b}} & \texttt{Isometric} & 1.18 & 28.25 & 84.97 & 35.01 & 1.18 & 31.20 & 87.59 & 43.92 & 1.10 & 48.70 & 85.64 & 37.12 \\
 & \texttt{Same} & 1.17 & 28.95 & 85.31 & 35.16 & 1.18 & 32.25 & 87.66 & 44.31 & 1.11 & 48.55 & 85.63 & 36.80 \\
 & \texttt{Short} & 1.06 & 39.75 & 83.12 & 29.42 & 1.03 & 42.10 & 84.60 & 35.92 & 1.01 & 50.25 & 84.71 & 35.11 \\
 & \texttt{Tiny} & 0.95 & 34.90 & 80.71 & 23.33 & 0.90 & 37.25 & 81.08 & 26.91 & 0.83 & 37.00 & 81.01 & 25.22 \\
\cline{1-14}
 & \texttt{Random} & 1.13 & 32.40 & 81.50 & 26.26 & 1.13 & 35.20 & 83.19 & 33.89 & 1.06 & 48.35 & 82.46 & 30.69 \\
\multirow[c]{3}{*}{\texttt{llama3:8b}} & \texttt{Isometric} & 1.13 & 34.00 & 81.80 & 26.70 & 1.13 & 36.25 & 83.53 & 34.10 & 1.06 & 47.35 & 82.78 & 30.94 \\
 & \texttt{Same} & 1.14 & 34.25 & 81.96 & 26.40 & 1.14 & 34.15 & 83.86 & 34.56 & 1.07 & 51.05 & 83.11 & 31.16 \\
 & \texttt{Short} & 1.06 & 37.95 & 80.57 & 24.32 & 1.04 & 38.10 & 81.43 & 28.63 & 1.00 & 48.00 & 81.45 & 28.03 \\
 & \texttt{Tiny} & 1.00 & 37.10 & 79.11 & 20.94 & 0.97 & 37.05 & 79.69 & 26.43 & 0.90 & 42.25 & 79.13 & 23.72 \\
\cline{1-14}
 & \texttt{Random} & 1.15 & 33.30 & 80.57 & 23.37 & 1.17 & 36.20 & 82.27 & 30.12 & 1.07 & 49.60 & 82.34 & 28.71 \\
\multirow[c]{3}{*}{\texttt{mistral:7b}} & \texttt{Isometric} & 1.15 & 33.35 & 80.60 & 23.24 & 1.16 & 35.25 & 82.36 & 30.01 & 1.07 & 50.05 & 82.30 & 28.57 \\
 & \texttt{Same} & 1.16 & 33.50 & 80.90 & 23.88 & 1.17 & 35.40 & 82.55 & 29.78 & 1.09 & 49.50 & 82.82 & 29.06 \\
 & \texttt{Short} & 1.13 & 36.50 & 80.34 & 22.27 & 1.14 & 39.30 & 82.03 & 29.15 & 1.05 & 51.10 & 82.02 & 28.19 \\
 & \texttt{Tiny} & 1.12 & 39.00 & 79.55 & 21.25 & 1.15 & 39.95 & 80.96 & 28.10 & 1.04 & 51.20 & 81.77 & 27.70 \\
\cline{1-14}
 & \texttt{Random} & 1.42 & 29.35 & 82.14 & 30.68 & 1.30 & 32.20 & 84.55 & 38.96 & 1.44 & 41.80 & 83.27 & 33.30 \\
\multirow[c]{3}{*}{\texttt{mixtral:8x7b}} & \texttt{Isometric} & 1.35 & 32.15 & 82.24 & 31.23 & 1.24 & 33.65 & 84.60 & 38.34 & 1.30 & 44.55 & 83.83 & 34.02 \\
 & \texttt{Same} & 1.32 & 32.25 & 82.60 & 31.69 & 1.23 & 34.10 & 84.95 & 38.54 & 1.27 & 42.20 & 84.09 & 34.79 \\
 & \texttt{Short} & 1.33 & 35.65 & 81.56 & 29.40 & 1.22 & 36.85 & 83.80 & 35.92 & 1.27 & 46.90 & 83.64 & 33.99 \\
 & \texttt{Tiny} & 1.24 & 38.40 & 80.73 & 27.13 & 1.21 & 37.15 & 82.91 & 34.10 & 1.24 & 45.45 & 82.82 & 32.94 \\
\cline{1-14}
 & \texttt{Random} & 1.21 & 30.10 & 84.04 & 33.19 & 1.19 & 34.05 & 87.06 & 43.62 & 1.12 & 45.15 & 85.66 & 37.59 \\
\multirow[c]{3}{*}{\texttt{qwen2:72b}} & \texttt{Isometric} & 1.21 & 30.35 & 84.04 & 33.06 & 1.18 & 36.00 & 86.97 & 42.64 & 1.11 & 46.30 & 85.58 & 37.18 \\
 & \texttt{Same} & 1.21 & 31.10 & 84.06 & 33.12 & 1.17 & 35.15 & 87.14 & 43.29 & 1.12 & 48.60 & 85.65 & 37.30 \\
 & \texttt{Short} & 1.14 & 33.95 & 83.97 & 32.04 & 1.17 & 41.50 & 86.24 & 40.66 & 1.11 & 48.90 & 85.38 & 36.94 \\
 & \texttt{Tiny} & 1.15 & 36.35 & 83.51 & 31.21 & 1.15 & 42.25 & 85.99 & 40.03 & 1.11 & 48.50 & 84.58 & 35.28 \\
\cline{1-14}
 & \texttt{Random} & 1.17 & 29.94 & 81.22 & 23.97 & 1.20 & 30.35 & 83.96 & 34.96 & 1.10 & 44.65 & 83.40 & 30.81 \\
\multirow[c]{3}{*}{\texttt{qwen2:7b}} & \texttt{Isometric} & 1.17 & 31.15 & 81.27 & 23.54 & 1.19 & 30.45 & 84.01 & 35.03 & 1.10 & 46.35 & 83.60 & 31.22 \\
 & \texttt{Same} & 1.18 & 30.15 & 81.39 & 24.22 & 1.19 & 33.05 & 84.43 & 35.24 & 1.10 & 44.70 & 83.35 & 30.91 \\
 & \texttt{Short} & 1.15 & 33.85 & 80.92 & 23.33 & 1.17 & 33.85 & 83.48 & 34.25 & 1.09 & 46.80 & 83.43 & 30.65 \\
 & \texttt{Tiny} & 1.14 & 34.25 & 80.34 & 21.87 & 1.15 & 34.85 & 82.98 & 33.09 & 1.07 & 49.50 & 83.11 & 30.23 \\\hline
 Oracle &  & 1.07 & 65.00 & 87.74 & 49.60 & 1.05 & 76.50 & 87.99 & 55.60 & 1.03 & 83.00 & 88.47 & 53.50 \\\hline
\end{tabular}
    \caption{20-shot prompting for all language pairs when sampling examples from varying pools. We report length ratio (LR), length compliance (LC), BERTScore (BS), and BLEU. All numbers are averaged across 10 instances. The prompt text \emph{matches} the pool type. }
    \label{tab:pool_type_evaluation_20}
\end{table*}

\end{document}